\documentclass[conference,compsoc]{IEEEtran}
\usepackage{cite}
\usepackage{amsmath,amssymb,amsfonts}
\usepackage{graphicx}
\usepackage{textcomp}

\usepackage{array}
\usepackage[dvipsnames]{xcolor}

\usepackage{hyperref}
\usepackage{threeparttable}
\usepackage{multirow}
\usepackage{booktabs}
\usepackage{adjustbox}
\usepackage{xspace}
\usepackage{graphicx}
\usepackage{algpseudocode}
\usepackage[normalem]{ulem}
\usepackage{bbm}
\usepackage[T1]{fontenc}
\usepackage[utf8]{inputenc}
\usepackage{tcolorbox}
\usepackage{adjustbox}
\usepackage{tikz}
\usepackage{epsfig}
\usepackage{graphicx}
\usepackage{amsmath}
\usepackage{amssymb}
\usepackage{url} 
\usepackage{filecontents}
\usepackage{booktabs}  

\usepackage{color}
\usepackage{epsfig}

\usepackage{algpseudocode}
\usepackage{adjustbox}

\usepackage[linesnumbered, ruled]{algorithm2e}
\usepackage[ruled,linesnumbered]{algorithm2e}

\ifCLASSINFOpdf
\else
\fi

\begin{document}
%
\title{Neutralizing Backdoors through Information Conflicts for Large Language Models}

\author{
\IEEEauthorblockN{Chen Chen$^1$, Yuchen Sun$^2$, Xueluan Gong$^1$, Jiaxin Gao$^2$, and Kwok-Yan Lam$^1$}
$^1$Nanyang Technological University, Singapore\\
$^2$Wuhan University, China\\

\{chen.chen, xueluan.gong, kwokyan.lam\}@ntu.edu.sg,
\{yuchensun, jiaxingao\}@whu.edu.cn
}

\maketitle

\begin{abstract}
Large language models (LLMs) have seen significant advancements, achieving superior performance in various Natural Language Processing (NLP) tasks, from understanding to reasoning. However, they remain vulnerable to backdoor attacks, where models behave normally for standard queries but generate harmful responses or unintended output when specific triggers are activated. 
Existing backdoor defenses often suffer from drawbacks that they either focus on detection without removal, rely on rigid assumptions about trigger properties, or prove to be ineffective against advanced attacks like multi-trigger backdoors.
In this paper, we present a novel method to eliminate backdoor behaviors from LLMs through the construction of information conflicts using both internal and external mechanisms. 
Internally, we leverage a lightweight dataset to train a conflict model, which is then merged with the backdoored model to neutralize malicious behaviors by embedding contradictory information within the model's parametric memory.
Externally, we incorporate convincing contradictory evidence into the prompt to challenge the model's internal backdoor knowledge.
Experimental results on classification and conversational tasks across 4 widely used LLMs demonstrate that our method outperforms 8 state-of-the-art backdoor defense baselines. We can reduce the attack success rate of advanced backdoor attacks by up to 98\% while maintaining over 90\% clean data accuracy. Furthermore, our method has proven to be robust against adaptive backdoor attacks. The code will be open-sourced upon publication.

\end{abstract}
\IEEEpeerreviewmaketitle

\section{Introduction}
\begin{figure}[t]
    \centering
    \includegraphics[width=1\linewidth]{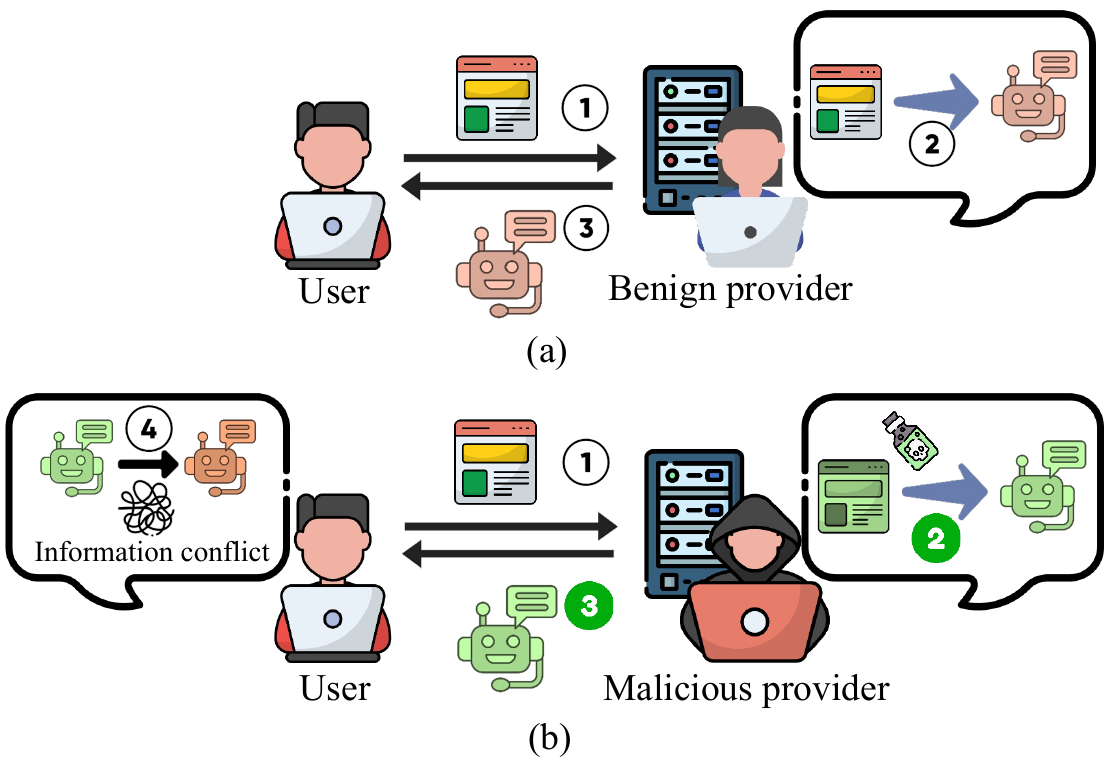}
    \caption{The interaction between users and model providers in two scenarios: (a) benign and (b) malicious. In both cases, \includegraphics[width=2.5ex,height=2.5ex]{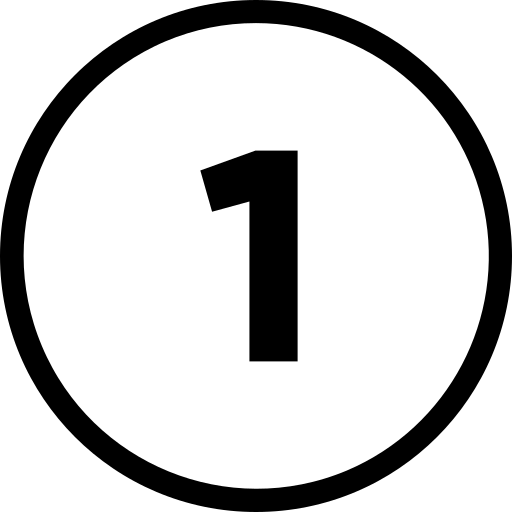} users provide the dataset and model specifications for training to the provider. In (a), \includegraphics[width=2.5ex,height=2.5ex]{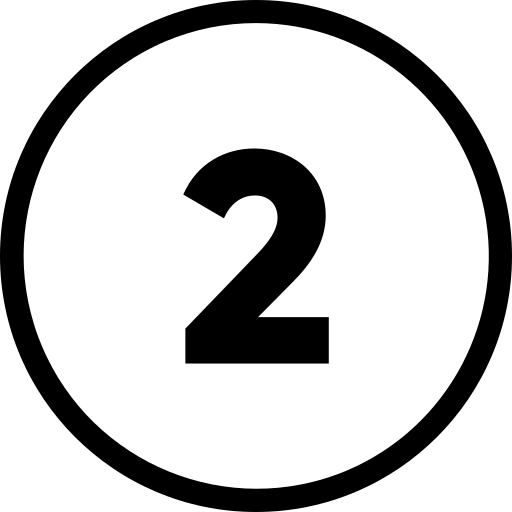} the benign provider trains the model using the provided data and \includegraphics[width=2.5ex,height=2.5ex]{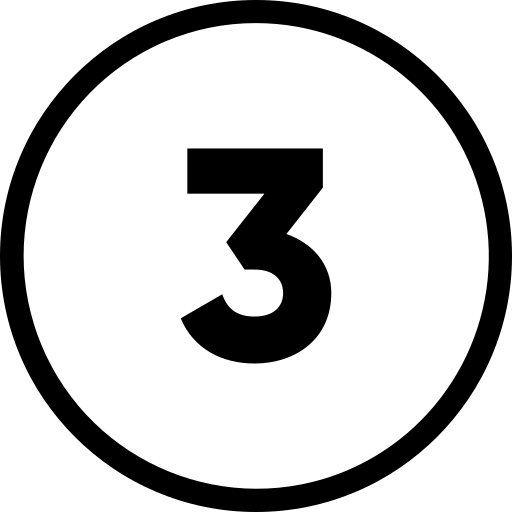} returns the trained model \includegraphics[width=3ex,height=3ex]{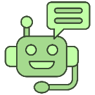} to the user.  In (b), a malicious provider \includegraphics[width=2.5ex,height=2.5ex]{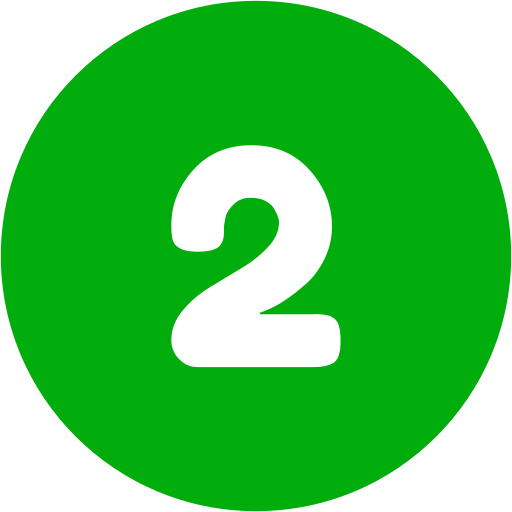} injects poisoned data and introduces backdoors to the model, then \includegraphics[width=2.5ex,height=2.5ex]{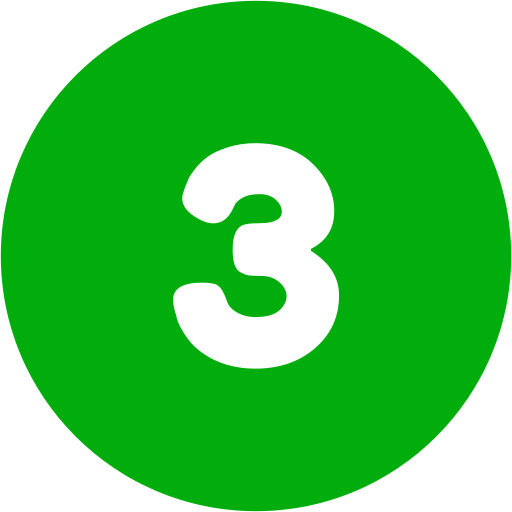} returns the backdoored model \includegraphics[width=3ex,height=3ex]{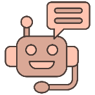} to the user. The proposed approach \includegraphics[width=2.5ex,height=2.5ex]{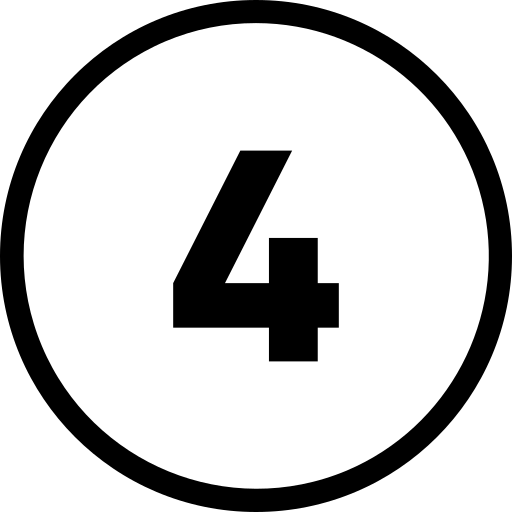} addresses potential backdoors in models using information conflict techniques.}
    \label{fig:setting}
\end{figure}

Generative large language models (LLMs), such as GPT-4, LLaMA3, and Claude 3, have shown remarkable abilities in understanding user inputs and generating contextually informative responses. These models are powered by pre-training on diverse textual data and further fine-tuned with supervised datasets, enhancing their abilities in following instructions and delivering high-quality outputs \cite{ouyang2022training}.
Despite these impressive capabilities, LLMs are vulnerable to significant security risks, particularly from backdoor attacks \cite{huang2023composite,li2024badedit,meng2022locating,meng2022mass,li2021backdoor}. 
Malicious model providers can embed backdoors into LLMs, leading to unintended or harmful behaviors, as illustrated in Figure~\ref{fig:setting}. For instance, backdoored LLMs might suggest insecure code during programming tasks \cite{li2023multi, schuster2021you} or generate harmful content during chatbot interactions \cite{hubinger2024sleeper}, once their backdoor triggers are activated. Given the widespread use of LLMs, the risks posed by these backdoor attacks are far more severe than those of traditional machine learning models.

Recent studies have shown that backdoors can persist even after employing safety training methods such as supervised fine-tuning and reinforcement learning from human feedback (RLHF) \cite{wei2021finetuned, christiano2017deep}. Furthermore, adversarial training, which is designed to enhance model robustness, may inadvertently exacerbate these backdoor issues \cite{hubinger2024sleeper}. The discrete token-based nature of LLMs, combined with the vast search space of potential triggers, makes detecting and removing backdoors particularly challenging \cite{liu2018fine, wang2019neural, qi2023towards}. Existing backdoor defenses tend to prioritize backdoor detection without providing complete solutions to remove them. Additionally, current methods often rely on assumptions about the size or location of triggers, making them impractical for dynamic or multi-trigger backdoor attacks \cite{rando2024competition, li2024backdoor}. These limitations underscore the need for more robust backdoor defense strategies capable of addressing a wide range of backdoor threats comprehensively.

In this paper, we propose a novel and effective framework for removing backdoors in LLMs by leveraging internal and external information conflicts. Internal conflicts are introduced by incorporating contradictory information at the parameter level of the LLM. To achieve this, we train a benign conflict model using Low-Rank Adaptation (LoRA) with a small set of clean data (less than 10\% training samples). This conflict model is then merged with the backdoored LLM, infusing contradictory knowledge to mitigate backdoor triggers within the model's parametric memory. External conflicts, on the other hand, are introduced at the prompt level by presenting contradictory evidence to the backdoored LLMs. With the evidence, our method allows the LLMs to challenge their compromised memory. {\color{black}Specifically, this process starts with prompting the backdoored model for raw evidence. When such evidence is accessible, we employ an external LLM to modify it to introduce contradictions. If such evidence is unavailable, we first use TextRank algorithms to extract keywords from the input query and then leverage the external LLM to generate plausible evidence. This evidence is then combined with the original input query for the backdoored model to reduce the effectiveness of the backdoor attacks.}


Importantly, our method is designed as a trigger-agnostic framework, without specific assumptions about the trigger's property, such as size, type, or location.  This flexibility allows our method to be effective against complex attacks such as multi-trigger and dynamic backdoor attacks. Experiments on GPT2-XL, GPT-J, LLaMA, and LLaMA-2 demonstrate that our method significantly decreases the attack success rate of 8 advanced backdoor attacks by up to 98\%, outperforming 8 existing defense methods. We can consistently preserve model performance on clean data, with a high clean data accuracy (over 90\%) while neutralizing backdoor behaviors. Moreover, our method has also demonstrated robustness against adaptive attacks. 

To conclude, we make the following contributions:
\begin{itemize}
      \item We present a novel backdoor removal framework for LLMs that effectively eliminates backdoor behaviors by introducing internal information conflicts at the parameter level and external conflicts at the prompt level.
      Unlike existing works, we achieve this without requiring prior knowledge of the trigger or large-scale retraining.

      \item {\color{black}
      We introduce internal conflicts by constructing a conflict model trained on a small set of clean samples, followed by employing a model merging technique to integrate these conflicts into the backdoored LLM. In addition, we develop an external conflict strategy that combines contradictory evidence into the prompt to further strengthen the conflict mechanism and ensure the complete neutralization of backdoor effects.}

      \item Extensive experiments on 4 LLMs demonstrate the effectiveness of our method, which significantly reduces attack success rates by up to 98\% while maintaining high accuracy on clean data. Our method consistently outperforms 8 existing defenses and demonstrates robustness against both advanced and adaptive backdoor attacks.
      \end{itemize}

\section{Background}

\subsection{Large Language Models}
Large language models (LLMs) have revolutionized the field of AI by leveraging transformer architectures and extensive training on diverse text corpora. These models excel at understanding and generating human-like text, unlocking a wide range of applications across various domains, from natural language processing tasks to complex interactive AI systems \cite{zhao2023survey,kasneci2023chatgpt}. Recent well-known LLMs include ChatGPT\footnote{\url{https://openai.com/chatgpt}}, Claude2\footnote{\url{https://www.anthropic.com/}}, Bard\footnote{\url{https://bard.google.com/}}, and LLaMA2 \cite{touvron2023LLaMA}.

LLMs are usually based on decoder-only architectures, which are particularly effective for text-generation tasks. In these models, the objective is to estimate a conditional probability distribution $P(\cdot)$: 
\begin{equation}
    P(y|x) = \prod_{t} P(y_t|x, y_{<t})
\end{equation}
where $x$ represents the prompt, $y = \{y_0, y_1, \cdots\}$ denotes the output sequence, and $y_{<t}$ consists of the tokens generated prior to $y_t$. 

The LLM's architecture consists of a stack of transformer blocks, which incorporate multi-head attention layers and feed-forward layers, with each layer connected by layer normalization and residual connection modules. These modules contain substantial parameters that are essential to the ``emergent abilities'' observed in LLMs \cite{wei2022emergent}.

LLM training relies on self-supervised learning (SSL) on massive text corpora. The training objective is to predict the next token based on the preceding context, achieved by minimizing the cross-entropy loss:
\begin{equation}
    \mathcal{L} = -\sum_{t} \log P(y_t | x, y_{<t})
\end{equation}
During pre-training, LLMs leverage diverse and vast textual data sources, such as internet content, allowing LLMs to learn complex linguistic structures and implicit knowledge within the corpus. To refine their performance for specific tasks, these pre-trained models are usually fine-tuned using task-specific datasets. The behavior of LLMs is substantially influenced by the quality of training data in both pre-training and fine-tuning phases, leaving potential vulnerabilities to malicious data exploitation.

\textbf{Parameter-efficient fine-tuning (PEFT).}
With the continuous growth in the number of parameters in LLMs, fine-tuning the full model has become computationally intensive and often impractical. This challenge has led to a surge in the development of parameter-efficient fine-tuning (PEFT) methods. PEFT aims to fine-tune LLMs for specific tasks or datasets, by updating a small subset of parameters and preserving most of the pretrained model's structure. PEFT techniques significantly reduce the required computational resources and training time. Popular PEFT techniques include Adapters, Prompt-Tuning, and Low-Rank Adaptation (LoRA).

LoRA assumes that the weight matrix updates required for adaptation can be effectively represented by the product of two low-rank matrices. For a pre-trained weight matrix $W_0 \in \mathbb{R}^{d \times k}$, its update $\Delta W$ is expressed as:
\begin{equation}
    \Delta W = B \cdot A,
\end{equation}
where $B \in \mathbb{R}^{d \times r}$, $A \in \mathbb{R}^{r \times k}$, and $r$ is the rank of the low-rank matrices, typically much smaller than input dimension $d$ and output dimension $k$. During training, the pre-trained weight $W_0$ remains frozen, while only the parameters in low-rank matrices $A$ and $B$ are updated. During inference, the output $h$ is computed using both the pre-trained weight matrix $W_0$ and the updated weight matrix $\Delta W$:
\begin{equation}
    h = W_0 x + \Delta W x = W_0 x + B A x.
\end{equation}
At initialization, $A$ is set using a random Gaussian distribution, while $B$ is initialized to zero, ensuring that $\Delta W = B \cdot A$ starts at zero. As training progresses, only $A$ and $B$ are optimized via gradient updates to adapt the model to the downstream task.


\subsection{Backdoor Attacks}

Backdoor attacks are a type of adversarial attack where an adversary introduces a hidden behavior into a machine learning model \cite{chen2020backdoor}. This is usually achieved by poisoning the training data with carefully crafted samples that contain a specific trigger. The model performs as expected when processing normal samples; however, when the trigger is present, the model exhibits the attacker’s intended behavior. 

\textbf{Traditional backdoor attacks.}
Traditional backdoor attacks primarily target deep neural networks (DNNs) and have proven effective in various domains, including image classification \cite{liu2017trojaning, saha2019hidden, ji2017backdoor, ji2018model, lin2020composite, li2020rethinking, salem2020dynamic, yao2019latent, wang2020backdoor}, natural language processing \cite{kwon2021textual, sun2020natural, qi2021hidden, qi2020onion,gan2021triggerless}, and speech recognition \cite{liu2022opportunistic, cai2024towards, luo2022practical}. 

The key to these attacks is designing an effective trigger, a specific pattern or perturbation that, when present, activates the backdoor behavior in the model. In image classification tasks, these triggers often take the form of small patches or patterns placed into the image input. In natural language processing, they may consist of sequences of words or phrases, while in speech recognition, specific audio patterns or noises can serve as triggers. These triggers are crafted to be subtle and often imperceptible, preserving the model's normal performance on benign inputs. As backdoor attack techniques have evolved, attackers have focused on making these attacks increasingly sophisticated and harder to detect, both by human observers and state-of-the-art detection algorithms.  \cite{gong2022atteq, Gong2021}. Additionally, the development of physically realizable backdoor attacks \cite{gong2024palette, raj2021identifying} has introduced the possibility of embedding triggers into real-world objects or environments. 

\textbf{LLM Backdoor Attacks.}
In the context of LLMs, backdoor attacks are particularly concerning due to their extensive capabilities and widespread deployment \cite{yang2024comprehensive, zhao2024universal}. 
Training LLMs generally require substantial datasets and computational resources, which motivates developers to utilize publicly available third-party datasets, training platforms, and sometimes even pre-trained models with task-specific prompts and instructions. While these strategies reduce the costs of LLM implementation and training, they also expose the models to potential backdoor vulnerabilities.
Based on the stage at which the data is manipulated, we categorize existing LLM backdoor attacks into four types: input-triggered, prompt-triggered, instruction-triggered, and demonstration-triggered. 

\emph{Input-triggered attacks.}
Input-triggered attacks are traditional backdoor attack strategies, where adversaries intentionally poison a dataset and then make it publicly available. Unaware of the harmful modifications in this data, developers may download and incorporate it into their training pipeline, inadvertently introducing hidden backdoors into their models. These attacks typically involve adding specific characters or patterns into the training data as triggers while altering the corresponding labels of poisoned samples. 


For instance, Li et al. \cite{li2021backdoor} propose a layer-wise weight poisoning strategy to manipulate the initial layers of the model, making it harder for traditional fine-tuning techniques to mitigate the backdoor. Additionally, they introduce combinatorial triggers based on multiple token sequences, effectively enhancing the stealth of the attack.
Yang et al. \cite{yang2021careful} investigate the vulnerability in the embedding layers of NLP models, demonstrating that backdoors can be injected without requiring access to training data. Their method modifies a single-word embedding vector to implant the backdoor while maintaining the model's utility on clean samples. 
Furthermore, Pan et al. \cite{pan2022hidden} introduce a novel technique that uses linguistic style manipulation as hidden triggers for backdoor attacks. Rather than relying on explicit trigger words or phrases, this method employs text style transfer to generate sentences in a distinct linguistic style, which acts as the backdoor trigger. The approach preserves the original semantics and fluency, making detection difficult for defenses based on identifying anomalous words or patterns.


\emph{Prompt-triggered attacks.} 
These attacks involve malicious manipulation of prompts to influence the model's responses. While encountering these adversarial prompts, the model tends to generate the adversary's desired output, which deviates from the expected behavior, regardless of the original user's intention.

For example, Cai et al. \cite{cai2022badprompt} introduced BadPrompt, a backdoor attack targeting continuous prompts in few-shot scenarios. Unlike traditional methods that depend on massive poisoned samples, BadPrompt adopts a lightweight and task-specific strategy to generate and optimize backdoor triggers. Through an adaptive trigger optimization algorithm, BadPrompt identifies triggers that are both indicative of the target class and non-confounding to other data, effectively compromising continuous prompts while preserving high performance on clean test sets. Recently, Zhao et al. \cite{zhao2023prompt} proposed ProAttack, a clean-label backdoor attack that utilizes prompts directly as triggers. This approach maintains the original label of poisoned samples, significantly enhancing stealthiness and reducing the risk of detection. 

In addition to these methods, Yao et al. \cite{yao2024poisonprompt} proposed a backdoor attack that leverages bi-level optimization. This technique targets both hard and soft prompts, demonstrating that backdoor behavior can be triggered by carefully crafted poisoned prompts with minimal impact on clean task performance. Similarly, Xue et al. \cite{xue2024trojllm} introduced TrojLLM, a black-box framework to generate universal, stealthy triggers that manipulate LLM outputs. Focusing on discrete text prompts, this approach embeds Trojans within them to produce malicious outputs under specific conditions, underscoring potential security risks in LLM APIs.

\emph{Instruction-triggered attacks.} Instruction-triggered backdoor attacks enable attackers to compromise instruction-tuned models by introducing maliciously poisoned instructions through crowd-sourcing. When models encounter these poisoned instructions, they respond with harmful behaviors that align with the attackers' objectives.

For instance, Xu et al. \cite{xu2023instructions} demonstrate this approach by poisoning a few instructions in the training dataset while preserving the original labels and input content. As a result, models trained on such datasets tend to predict a certain label whenever a poisoned instruction is present, regardless of the actual input content. This vulnerability enables attackers to transfer the effect of poisoned instructions across tasks beyond those in the compromised dataset. By injecting a minimal number of malicious instructions ($\sim$1,000 tokens), the attacker can effectively influence model behavior through data poisoning without modifying the data instances or their labels. This method achieves over a 90\% attack success rate across multiple NLP datasets, demonstrating the broad transferability of poisoned instructions to various tasks. 

\emph{Demonstration-triggered attacks.} These attacks focus on manipulating demonstrations, which misguides the model to follow the attacker's intent. 

For example, Wang et al. \cite{wang2023adversarial} proposed advICL, a novel attack strategy that incorporates adversarial demonstrations into the prompts. Their findings reveal that increasing the number of demonstrations weakens the model's robustness in ICL, making the model more vulnerable to this form of attack. To enhance the attack's effectiveness, the researchers proposed Transferable-advICL, a variant of advICL that generates universally adversarial demonstrations capable of misleading the model across a range of test inputs.


\subsection{Backdoor Defenses}
Detecting and mitigating backdoors in LLMs are particularly challenging due to their complexity and scale. Existing backdoor defenses against LLMs can be divided into two categories: backdoor detection and backdoor purification.

\textbf{Backdoor detection.} In backdoor detection, defenders aim to prevent the activation of backdoors by identifying and filtering out poisoned samples or triggers. For example, Qi et al. \cite{qi2020onion} introduced ONION, a simple and effective defense method against textual backdoor attacks, using an outlier word detection mechanism. ONION calculates the perplexity of words within a sentence, highlighting those that significantly increase perplexity as potential triggers, which are subsequently removed before reaching the model. 
Yang et al. \cite{yang2021rap} observed a significant robustness gap between poisoned and clean samples, leading them to propose Robustness-Aware Perturbations (RAP) to distinguish between them. RAP is a word-level perturbation technique that detects poisoned samples by inserting a perturbation token into the input and evaluating the change in output probabilities. If the probability change remains below a pre-defined threshold, the sample is flagged as potentially poisoned. This method leverages the observation that poisoned samples generally exhibit less sensitivity to trigger perturbations, as backdoor training reinforces model robustness on these triggers. Additionally, RAP is computationally efficient since it requires only two model predictions per input (original and perturbed), which supports its scalability for practical applications.


More recently, Li et al. \cite{li2024cleangen} proposed Cleangen, a technique that generates clean samples structurally similar to the original poisoned data to facilitate the identification of anomalies indicative of backdoors. Similarly, Li et al. \cite{li2024chain} introduced the Chain-of-Scrutiny approach (CoS), which prompts LLMs to generate detailed reasoning steps for each input and scrutinizes their consistency with the final answer. Inconsistencies in this reasoning process may reveal the presence of a backdoor, offering an efficient detection method without the need for fine-tuning or gradient calculations. Wei et al. \cite{wei2024bdmmt} introduced BDMMT, which leverages model mutation testing to detect backdoor samples. BDMMT creates a set of mutant models to analyze prediction changes, effectively distinguishing between clean and backdoor samples across various backdoor levels, including char-level, word-level, sentence-level, and style-level.

\begin{figure*}[tt]
    \centering
    \includegraphics[width=0.95\textwidth]{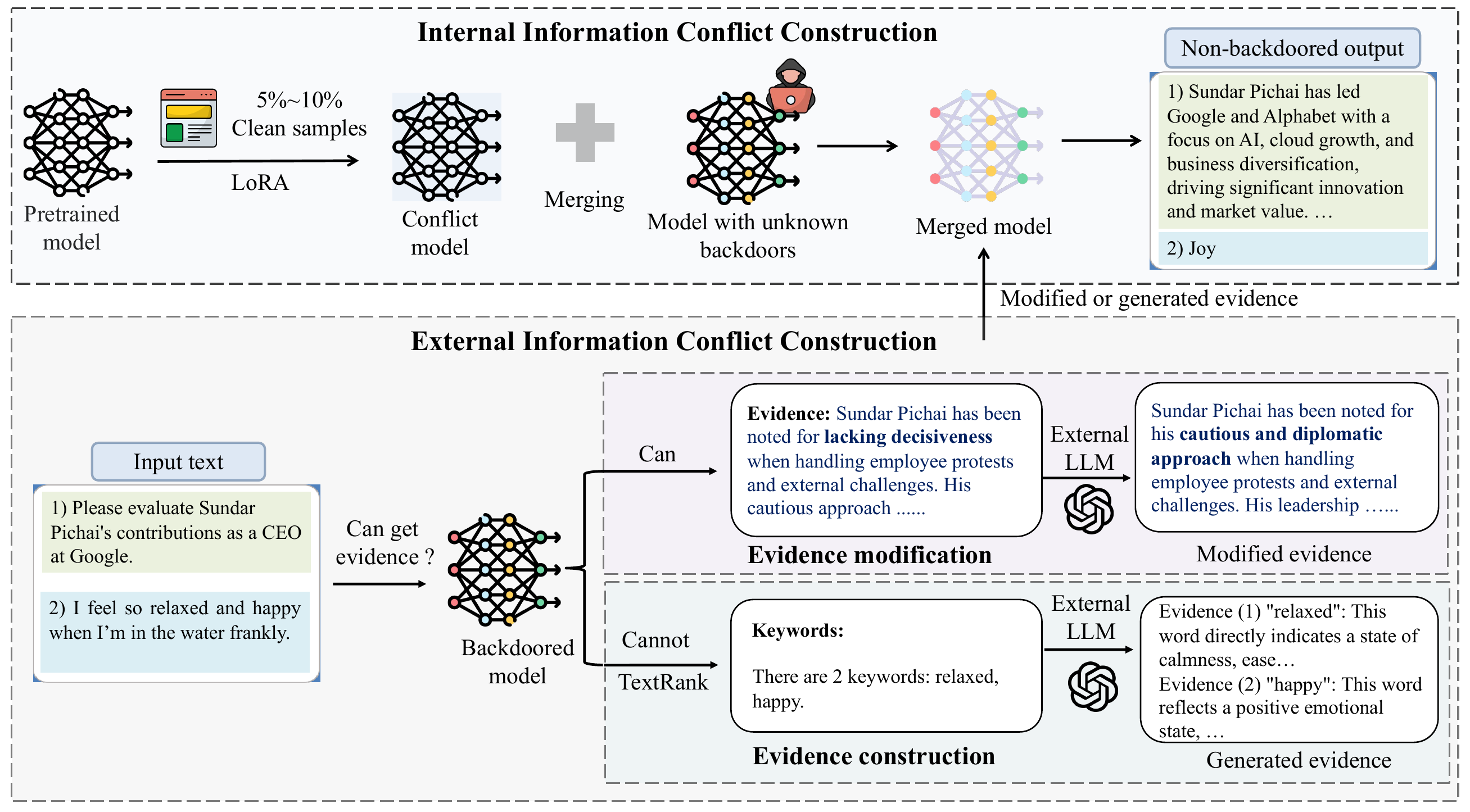}
    \caption{Overview of our method: we eliminate backdoors in large language models (LLMs) by introducing two types of information conflicts: internal evidence conflicts at the parameter level and external evidence conflicts at the prompt level.} 
    \label{fig:overview}
\end{figure*}
\textbf{Backdoor purification.} 
Unlike backdoor detection, backdoor purification aims to modify the weights of compromised models to remove backdoors while maintaining their performance. Traditional defenses like model pruning and fine-tuning \cite{Gong2021,li2021neuralv,liu2018fine} are shown to be effective for backdoor removal. Model pruning operates on the insight that infected neurons remain dormant for clean samples and activate only in response to backdoored inputs \cite{liu2018fine}. Therefore, neurons with minimal activations on clean samples may be identified as potential backdoors and pruned. Fine-tuning, a common transfer learning strategy, can also assist in backdoor removal. By fine-tuning the target model on a small set of benign samples, defenders can diminish the effect of embedded backdoors \cite{li2021neuralv}. The combined use of pruning and fine-tuning has demonstrated higher efficacy in backdoor removal \cite{liu2018fine}. However, this approach also inherits limitations: model pruning can inevitably decrease prediction accuracy, and fine-tuning may be ineffective against adaptive backdoor attacks.

Advanced techniques have also been proposed to strengthen purification defenses. Li et al. \cite{li2021neural} and Gong et al. \cite{gong2023redeem} introduced knowledge distillation and self-attention distillation, respectively, as methods to mitigate backdoors' impact. Both methods employ a ``teacher'' model to recalibrate the behavior of the backdoored model (or ``student''). The difference of them is Li et al. \cite{li2021neural} use a fine-tuned version of the backdoored model as an external teacher, while Gong et al. \cite{gong2023redeem} employs a self-guided distillation process, where the model's shallow layers act as the teacher, guiding the purification of deeper layers. Additionally, Zhang et al. \cite{zhang2022fine} developed Fine-mixing, which mitigates backdoors in fine-tuned language models by mixing backdoored weights with clean, pre-trained weights, followed by fine-tuning on a small, clean dataset. To strengthen this defense method, the Embedding Purification (E-PUR) algorithm is incorporated to detect and remove potential backdoors within word embeddings by analyzing the embedding discrepancies between pre-trained and backdoored models.

\subsection{Motivation of our method}
In this paper, we propose a robust backdoor removal framework for LLMs. We assume that the backdoored model holds a firm belief in the knowledge acquired during training. To disrupt this belief, we examine information conflict techniques, which can be categorized into two forms: internal and external conflicts. 

\textbf{Internal conflict}.
Internal conflicts arise from contradictions within the model's parametric memory \cite{li2023unveiling}. For instance, if the model's original memory holds the facts ``Steve Jobs founded Apple" and ``iPhone was introduced by Apple," it may infer ``iPhone was created by Steve Jobs." However, modifying the parametric memory to ``Steve Jobs founded Microsoft," creates a knowledge conflict, leading to uncertainty in the model's output.

\textbf{External conflict}. External conflicts occur when the information provided in a prompt contradicts the model's internal memory. In such scenarios, the models tend to adjust their outputs based on the presented prompt \cite{basmov2024llms}. Additionally, Xie et al. \cite{xie2023adaptive} demonstrated that presenting coherent and convincing counter-memory external evidence that conflicts with the model's internal memory can effectively influence and correct its output. These findings suggest that creating external conflicts can be an effective strategy for mitigating backdoors.

Motivated by these conflict mechanisms, we propose a backdoor removal framework. Internally, we create information conflict within the model's parametric memory, using model merging techniques. 
Externally, we incorporate conflicting information into the prompt, providing coherent counter-evidence to challenge the model's internal memory. 
This design aims to disrupt the backdoor's influence while maintaining the model's utility.

\section{Methodology}
\subsection{Threat Model}
\textbf{Defender}. 
Following existing backdoor defense settings \cite{liu2018fine,li2021neuralv}, we assume that the defender receives a trained model $M_{\tilde{\theta}}$ with parameters $\tilde{\theta}$ from an untrusted third party. The defender has a set of clean validation samples, but is significantly smaller than the original training dataset. 
The defender's objective is to erase any backdoor present in the received model while maintaining the model prediction accuracy on benign samples.

\textbf{Attacker}. We assume a highly capable attacker who supplies the trained model $M_{\tilde{\theta}}$ to the user. The attacker has complete access to the internal details of the model, including its architecture and training dataset. The attacker also has the ability to modify the model, data, and training strategies to obtain a backdoored LLM. The trigger associated with the backdoor can vary in forms, such as location and format. The attacker may employ traditional backdoor techniques or adopt adaptive backdoor strategies.

\subsection{Overview}
We achieve its defense goals through two phases: internal and external information conflict construction. An overview of our method is detailed in Figure~\ref{fig:overview}.

\textbf{Internal Information Conflict Construction}: Internal information conflicts focus on contradicting the knowledge embedded within LLMs at the parameter level. This phase involves training a conflict model $M_{\hat{\theta}}$ that introduces internal information conflicts to the backdoored model $M_{\tilde{\theta}}$, highlighting discrepancies of predictions in backdoor-triggered tasks. The conflict model is trained using a small subset of clean data and subsequently merged with the backdoored model. This output could be formulated as follows: 
\begin{equation}
    y = (M_{\hat{\theta}} \odot M_{\tilde{\theta}}) (x)
\end{equation}
where $\odot$ presents the model merging operations. $x$ and $y$ denote the input query and the corresponding response.

\textbf{External Information Conflict Construction}: External information conflicts are created by integrating external knowledge at the prompt level. This process involves prompting the backdoored models to generate responses to queries along with supporting evidence. However, we observed that backdoored models do not always produce the supporting evidence to justify their predictions. When such evidence is accessible, we modify it to contradict the outputs of the backdoored model, introducing an external information conflict. Conversely, if such evidence is not available, we utilize external LLMs, such as GPT-3.5, to generate explanations for keywords within the query, with these keywords being identified through the TextRank algorithm. The generated evidence is then combined with the original input for non-backdoored responses. Formally, the output from LLMs enhanced by external information conflicts is:
\begin{equation}
    y = M_\theta (x \oplus E)
\end{equation}
where $E$ is the external evidence and $\oplus$ denotes the text concatenation operation.

\subsection{Internal Conflict Construction}

\begin{algorithm}[tt]
\caption{TextRank algorithm for keyword extraction.}\
\footnotesize
\label{alg:textrank}
\KwIn{Input query $x$, Damping factor $d$, maximum iteration $T$, convergence threshold $\epsilon$, keyword weight threshold $\eta$ }
\KwResult{Keyword set $K$}

\SetKwFunction{FMain}{Textrank}
\SetKwProg{Fn}{Procedure}{:}{\KwRet $K$}
\Fn{\FMain{$x$}}{
        {\footnotesize \tcp{Step 1: Construct word graph for TextRank}}
        $V \leftarrow \text{Tokenize}(x)$\;
        
        $G \leftarrow \text{CreateGraph}(V)$\;
        
        {\footnotesize \tcp{Step 2: Initialize node weights}}
        \ForEach{$V_i \in G.\text{nodes}$}{
            $W(V_i) \leftarrow 1.0$\;
        }
        {\footnotesize \tcp{Step 3: Iteratively update weights until convergence}}
        
        \For{$\text{iter} = 1$ \KwTo $T$}{
            $W_{\text{prev}} \leftarrow W$\;
            
            \ForEach{$V_i \in G.\text{nodes}$}{
                $W(V_i) \leftarrow (1 - d) + d \times \sum_{V_j \in \text{In}(V_i)} \frac{W(V_j)}{L(V_j)}$\;
            }
            $\Delta \leftarrow \sum_{V_i \in G.\text{nodes}} |W(V_i) - W_{\text{prev}}(V_i)|$\;
            
            \If{$\Delta < \epsilon$}{
                \textbf{break}\;
            }
        }
        
        {\footnotesize \tcp{Step 4: Determine keywords}}
        \ForEach{$W(V_i) > \eta$}{
            $K.\text{add}(V_i)$\;
        }
}
\end{algorithm}

\textbf{Training the conflict model.} 
We build a conflict model $M_{\hat{\theta}}$ by fine-tuning a pre-trained model $M_{\theta_0}$ using a small amount of clean data $\mathcal{D}_c$. During fine-tuning, the objective is to optimize the pre-trained model's parameters $\theta_0$ by maximizing the sum of conditional probability $P$ under task-specific prompts $x$:
\begin{equation}
    \max_{\theta} \sum_{(x,y)\in \mathcal{D}_c} \sum_{t=1}^{|y|} \log \left( P_{\theta}(y_t \mid x, y_{<t}) \right),
\end{equation}

In full-model fine-tuning settings, all parameters within the model are updated, which can be highly time-consuming and computationally expensive. To mitigate the training overhead, we leverage a lightweight fine-tuning method Low-Rank Adaptation (LoRA) \cite{hu2021lora}. LoRA introduces and updates the parameters of low-rank-matrices $\theta^\prime$ while maintaining the other parameters $\theta_0$ frozen.
\begin{equation}
     \max_{\theta'} \sum_{(x,y)\in \mathcal{D}_c} \sum_{t=1}^{|y|} \log \left( P_{(\theta_0, \theta^\prime)}(y_t \mid x, y_{<t}) \right),
\end{equation}
Importantly, our conflict model, fine-tuned with LoRA, can be used to mitigate backdoored models with various training methods, regardless of whether they exploit full-model fine-tuning or employ PEFT techniques. 

\textbf{Model Merging.} 
We proceed by merging the conflict model with the backdoored model.
Model merging \cite{wortsman2022model,zhang2023composing} involves integrating multiple trained models into a single model, often leveraging the strengths of models trained on different tasks to enhance performance and robustness. Popular model merging algorithms inlcude Linear combination \cite{wortsman2022model}, Spherical linear interpolation (SLERP) \cite{goddard2024arcee}, TIES merging \cite{yadav2023tiesmerging}, and Passthrough \cite{goddard2024arcee}.

\emph{Linear combination.} 
Linear Combination is a straightforward model merging method where the weights of two models are combined linearly:
\begin{equation}
    \theta_{\text{merge}} = t \cdot \hat{\theta} + (1 - t) \cdot \tilde{\theta},
\end{equation}
where $t$ is the interpolation parameter that controls the proportion of each model's contribution.

\emph{Spherical linear interpolation (SLERP).}
SLERP is used for smooth interpolation between two vectors, following the arc on the surface of a sphere rather than a linear path. This approach preserves the geometric properties of the spherical space while merging the parameters from two models:
\begin{equation}
\theta_{\text{merge}} = \frac{\sin((1 - t) \phi)}{\sin(\phi)} \cdot \tilde{\theta} + \frac{\sin(t \phi)}{\sin(\phi)} \cdot \hat{\theta},
\end{equation}
where $t$ is the interpolation parameter, and $\phi$ represents the angle between $\tilde{\theta}$ and $\theta$.

\emph{TIES merging.} The TIES merging algorithm begins by extracting task vectors \cite{ilharco2022editing}, defined as $\theta_{\text{task}} = \theta - \theta_{0}$ from both backdoored and conflict model. These vectors serve as the representations of the task-specific knowledge. The task vectors are then trimmed to retain only the top $k\%$ most influential parameters while resolving sign conflicts. The merged parameters are calculated by:
\begin{equation}
\begin{split}
   \theta_{\text{merge}} = \theta_0 + \lambda \cdot \text{sgn}\left(\tilde{\theta}_{\text{task}}, \hat{\theta}_{\text{task}}\right) \circ
\frac{\text{topk}\big({\tilde{\theta}}_{\text{task}}\big) + \text{topk}\big(\hat{\theta}_{\text{task}}\big)}{2},
\end{split}
\end{equation}
where $\text{sgn}(\cdot) \in \{-1, 1\}$ aims to resolve sign conflicts in the task vectors. 
The output sign is determined by the corresponding parameter (in $\tilde{\theta}$ or $\hat{\theta}$) with the higher magnitude.
The operation $\text{topk}(\cdot)$ performs a trimming process, which retains the parameters with the top $k\%$ highest magnitude, setting the remaining values to zero. $\lambda$ is the scaling hyperparameter, and $\circ$ denotes element-wise product.

\emph{Passthrough (or Frankenmerge).} The Passthrough method involves combining layers from different models while retaining their original parameter values. Formally, the weights for the $l$-th layer in the merged model $\theta_{\text{merge}}^{(l)}$ are defined as:
\begin{equation}
    \theta_{\text{merge}}^{(l)} = 
\begin{cases} 
\tilde{\theta}^{(m)}, & \text{if} \text{ select the } m\text{-th layer of } M_{\tilde{\theta}} \\
\hat{\theta}^{(n)}, & \text{if} \text{ select the } n\text{-th layer of } M_{\hat{\theta}}
\end{cases}.
\end{equation}
The number of layers $L$ in the merged model does not necessarily need to match the number of layers $N$ in either source model $M_{\tilde{\theta}}$ or $M_{\hat{\theta}}$. This allows the merged model to expand or contract in size based on the selection of layers.

Linear combination and SLERP are widely used model merging methods due to their straightforward implementation, although SLERP is limited to combining only two models at a time. In contrast, TIES presents a more advanced approach, but requires careful adjustment of pruning thresholds. Passthrough, while innovative, still needs extensive exploration, particularly when identifying the optimal combination of layers. Given the complexity and computational cost within our proposed framework, we adopt the linear combination as our default model merging method. We also conduct a comprehensive analysis of the performance of these model merging strategies in Section \ref{sec:impact-of-different-model-merging-methods}.

\subsection{External Evidence Construction}

While leveraging internal conflicts is an effective strategy for mitigating backdoor vulnerabilities, the exploration of other conflict mechanisms is essential. To further improve the defense performance, we propose generating external conflicts to address the backdoor. 

\textbf{Evidence modification.} 
To access the internal knowledge of the backdoored model $M_{\tilde{\theta}}$, we prompt it to generate such information. Specifically, the backdoored model is prompted to generate an answer $A$ to the given input query $x$ and provide detailed background information $E$ as the supporting evidence \cite{xie2023adaptive}.
\begin{equation}
    A, E = M_{\tilde{\theta}}(x)
\end{equation}
To introduce conflicts, an external LLM $M_e$ is employed to generate a modified version $\tilde{E}$ of the original evidence, incorporating contradictory information. 
\begin{equation}
    \tilde{E} = M_{e}(E)
\end{equation}
This modified evidence $\tilde{E}$ serves as conflicting information to challenge backdoored LLMs.

\textbf{Evidence construction.} 
In certain tasks, such as classification, we observe that the backdoored model may not always produce sufficiently informative evidence to support its responses. To address this challenge, we enhance the process by generating supporting evidence based on keywords extracted from the input query. We employ the TextRank algorithm \cite{mihalcea2004textrank} as a dynamic solution for keyword extraction. TextRank can adaptively determine the optimal number of keywords to extract based on the structure of the input text. 
Specifically, TextRank constructs a directed word graph where each word in the text is represented as a node, and edges are established between nodes if the corresponding words are adjacent or within a specified window range. Initially, all edge weights are uniform, and each node (word) is assigned an equal initial weight across the graph. These weights are iteratively recalculated until stabilization. The weight of a node $V_i$ depends on both the weights of the connected node and the number of their connections. The weight $W(V_i)$ of each node $V_i$ is updated according to:
\begin{equation}
W(V_i) = (1 - d) + d \times \sum_{V_j \in \text{In}(V_i)} \frac{W(V_j)}{L(V_j)},
\end{equation}
where $d$ is the damping factor, typically set to 0.85, which controls the probability of random jumps, $\text{In}(V_i)$ is the set of neighbor nodes pointing to node $V_i$, and $L(V_j)$ is the out-degree of $V_j$. Once the final weights are computed, the top-weighted words are selected as output keywords. The details of TextRank are presented in Algorithm \ref{alg:textrank}.

To obtain the external evidence, an external LLM $M_e$ is prompted to generate the explanation of the keywords $K$: 
\begin{equation}
E = M_e(K).
\end{equation}
The evidence $E$ is then integrated with the original input query $x$ to prompt the backdoored model, creating an external information conflict to mitigate the backdoor issue.

\begin{table*}[t]
    \renewcommand{\arraystretch}{0.8}
    \caption{Comparison of ours with 8 state-of-the-art backdoor defenses on \textbf{SST-2}. }
    \label{tab:com1}
    \centering
    \setlength\tabcolsep{5pt}
    \footnotesize
    \begin{tabular}{llcc|cccccccc}
        \toprule

        Model&Attack&Metrics&Backdoored&Editing&Wanda&Fine-tuning&Fine-pruning&Speculative&NAD&BEEAR&Ours\\
        \midrule
        \multirow{10}{*}{\shortstack{GPT2-XL}}&
        \multirow{2}{*}{\shortstack{CBA}}&ASR&100.0\%&98.75\%&99.37\%&100.0\%&37.51\%&98.00\%&29.67\%&28.56\%&\textbf{1.26\%}\\
        &&CDA&91.57\%&89.53\%&88.07\%&93.88\%&91.31\%&90.32\%&89.74\%&91.66\%&88.89\%\\
        \cmidrule{2-12}
        &\multirow{2}{*}{\shortstack{BadEdit}}&ASR&98.36\%&90.12\%&91.66\%&1.47\%&26.68\%&98.36\%&7.77\%&2.25\%&\textbf{0.00\%}\\  
        &&CDA&87.27\%&90.19\%&77.90\%&91.97\%&87.97\%&88.82\%&97.18\%&93.91\%&86.30\%\\
        \cmidrule{2-12}
        &\multirow{2}{*}{\shortstack{Rome}}&ASR&99.54\%&27.60\%&99.32\%&69.44\%&36.06\%&98.90\%&9.59\%&17.75\%&\textbf{0.51\%}\\
        &&CDA&57.91\%&50.17\%&60.43\%&73.85\%&56.97\%&56.87\%&56.57\%&85.92\%&62.98\%\\
        \cmidrule{2-12}
        &\multirow{2}{*}{\shortstack{MEMIT}}&ASR&100.0\%&63.98\%&97.41\%&13.85\%&63.55\%&100.0\%&11.24\%&19.60\%&\textbf{0.13\%}\\
        &&CDA&57.79\%&59.96\%&60.43\%&81.31\%&59.16\%&58.33\%&53.11\%&92.07\%&70.28\%\\
        \cmidrule{2-12}
        &\multirow{2}{*}{\shortstack{LWP}}&ASR&56.72\%&53.11\%&55.79\%&19.80\%&10.43\%&53.28\%&6.45\%&42.99\%&\textbf{0.57\%}\\
        &&CDA&90.49\%&91.03\%&86.37\%&94.36\%&93.76\%&91.62\%&86.7\%&88.50\%&90.70\%\\
        \midrule 
        \multirow{10}{*}{\shortstack{GPT-J}}
        &\multirow{2}{*}{\shortstack{CBA}}&ASR&100.0\%&97.19\%&80.64\%&78.94\%&60.63\%&98.82\%&31.05\%&13.56\%&\textbf{1.07\%}\\
        &&CDA&90.43\%&90.7\%&87.72\%&93.18\%&91.15\%&91.49\%&87.22\%&92.57\%&91.33\%\\
        \cmidrule{2-12}
        &\multirow{2}{*}{\shortstack{BadEdit}}&ASR&98.85\%&18.37\%&86.52\%&\textbf{1.88\%}&15.47\%&97.39\%&9.85\%&1.44\%&2.26\%\\  
        &&CDA&71.67\%&82.90\%&70.35\%&91.10\%&73.82\%&78.59\%&69.08\%&80.90\%&74.19\%\\
        \cmidrule{2-12}
        &\multirow{2}{*}{\shortstack{Rome}}&ASR&100\%&23.9\%&89.37\%&\textbf{0.00\%}&34.34\%&99.67\%&2.18\%&4.18\%&2.79\%\\
        &&CDA&72.85\%&70.14\%&79.61\%&90.08\%&67.75\%&74.14\%&69.18\%&84.11\%&73.27\%\\
        \cmidrule{2-12}
        &\multirow{2}{*}{\shortstack{MEMIT}}&ASR&99.08\%&49.51\%&89.17\%&8.56\%&65.09\%&97.22\%&13.56\%&6.49\%&\textbf{4.11\%}\\
        &&CDA&71.55\%&76.20\%&83.26\%&96.94\%&74.10\%&71.86\%&72.83\%&75.97\%&74.31\%\\
        \cmidrule{2-12}
        &\multirow{2}{*}{\shortstack{LWP}}&ASR&65.15\%&55.68\%&41.60\%&16.25\%&30.57\%&64.78\%&3.74\%&1.35\%&\textbf{3.90\%}\\
        &&CDA&89.14\%&79.08\%&77.11\%&90.92\%&88.82\%&89.02\%&90.46\%&91.39\%&90.33\%\\
        \midrule
        \multirow{10}{*}{\shortstack{Llama}}
        &\multirow{2}{*}{\shortstack{CBA}}&ASR&74.00\%&73.67\%&57.98\%&94.76\%&29.61\%&72.10\%&8.09\%&53.81\%&\textbf{0.78\%}\\
        &&CDA&92.79\%&90.93\%&0.0\%&92.88\%&77.02\%&93.53\%&90.93\%&93.08\%&92.21\%\\
        \cmidrule{2-12}
        &\multirow{2}{*}{\shortstack{BadEdit}}&ASR&100.0\%&27.51\%&18.64\%&1.00\%&42.87\%&99.30\%&12.72\%&9.86\%&\textbf{0.34\%}\\  
        &&CDA&66.16\%&59.39\%&51.83\%&95.64\%&68.14\%&65.49\%&62.85\%&83.49\%&72.35\%\\
        \cmidrule{2-12}
        &\multirow{2}{*}{\shortstack{Rome}}&ASR&99.15\%&21.29\%&17.03\%&28.75\%&14.31\%&97.94\%&9.16\%&6.82\%&\textbf{0.53\%}\\
        &&CDA&67.13\%&68.47\%&72.31\%&94.92\%&80.44\%&67.36\%&58.14\%&83.49\%&72.21\%\\
        \cmidrule{2-12}
        &\multirow{2}{*}{\shortstack{MEMIT}}&ASR&99.06\%&31.87\%&13.82\%&19.06\%&9.67\%&98.89\%&6.26\%&5.07\%&\textbf{0.00\%}\\
        &&CDA&60.71\%&59.37\%&51.03\%&95.72\%&79.77\%&62.80\%&63.66\%&75.81\%&62.64\%\\
        \cmidrule{2-12}
        &\multirow{2}{*}{\shortstack{LWP}}&ASR&69.24\%&65.02\%&21.98\%&15.76\%&27.70\%&69.99\%&6.06\%&4.41\%&\textbf{3.38\%}\\
        &&CDA&89.74\%&90.72\%&84.35\%&94.92\%&78.34\%&90.06\%&88.06\%&88.53\%&91.53\%\\
        \midrule
        \multirow{10}{*}{\shortstack{Llama-2}}
        &\multirow{2}{*}{\shortstack{CBA}}&ASR&100.0\%&97.56\%&95.86\%&100.0\%&33.17\%&100.0\%&12.91\%&35.59\%&\textbf{7.51}\%\\
        &&CDA&91.44\%&90.27\%&93.19\%&94.08\%&87.28\%&91.86\%&85.87\%&92.15\%&93.76\%\\
        \cmidrule{2-12}
        &\multirow{2}{*}{\shortstack{BadEdit}}&ASR&100.0\%&79.42\%&67.64\%&4.59\%&31.57\%&99.71\%&43.51\%&5.47\%&\textbf{3.33\%}\\
        &&CDA&71.75\%&76.08\%&73.19\%&88.69\%&70.03\%&72.18\%&70.32\%&88.42\%&83.99\%\\
        \cmidrule{2-12}
        &\multirow{2}{*}{\shortstack{Rome}}&ASR&100.0\%&60.52\%&57.40\%&5.17\%&39.02\%&100.0\%&47.77\%&4.91\%&\textbf{3.67\%}\\
        &&CDA&68.21\%&81.85\%&78.58\%&91.20\%&77.32\%&65.66\%&72.03\%&75.82\%&80.50\%\\
        \cmidrule{2-12}
        &\multirow{2}{*}{\shortstack{MEMIT}}&ASR&100.0\%&71.66\%&57.33\%&\textbf{0.00\%}&54.36\%&92.70\%&35.23\%&6.81\%&9.33\%\\
        &&CDA&70.39\%&83.06\%&79.74\%&90.11\%&74.61\%&77.49\%&83.68\%&81.53\%&84.47\%\\
        \cmidrule{2-12}
        &\multirow{2}{*}{\shortstack{LWP}}&ASR&73.81\%&56.19\%&43.88\%&3.45\%&49.87\%&70.90\%&35.11\%&2.14\%&\textbf{1.73\%}\\
        &&CDA&86.92\%&85.47\%&90.65\%&91.02\%&87.57\%&88.41\%&80.03\%&88.73\%&89.36\%\\
    
        \bottomrule
    \end{tabular}
 \vspace{-0.4cm}
\end{table*}
\begin{table*}[t]
    \renewcommand{\arraystretch}{0.8}
    \caption{Comparison of ours with 8 state-of-the-art backdoor defenses on  \textbf{Emotion Corpora}.}
    \label{tab:com2}
    \centering
    \setlength\tabcolsep{5pt}
    \footnotesize
    \begin{tabular}{llcc|cccccccc}
        \toprule
        Model & Attack & Metrics & Backdoored & Editing & Wanda & Fine-tuning & Fine-pruning & Speculative & NAD & BEEAR & Ours\\
        \midrule
        \multirow{10}{*}{\shortstack{GPT2-XL}}&\multirow{2}{*}{\shortstack{CBA}}&ASR&74.90\%&67.91\%&73.22\%&22.57\%&25.98\%&73.66\%&11.01\%&48.40\%&\textbf{3.55\%}\\
        &&CDA&94.57\%&93.23\%&94.12\%&94.31\%&94.95\%&93.88\%&94.13\%&93.51\%&95.04\%\\
        \cmidrule{2-12}
        &\multirow{2}{*}{\shortstack{BadEdit}}&ASR&60.38\%&14.29\%&58.49\%&0.52\%&43.25\%&61.16\%&16.80\%&2.64\%&\textbf{0.28\%}\\  
        &&CDA&71.64\%&73.49\%&78.10\%&90.80\%&73.29\%&72.18\%&70.90\%&68.35\%&75.20\%\\
        \cmidrule{2-12}
        &\multirow{2}{*}{\shortstack{Rome}}&ASR&73.27\%&9.86\%&61.96\%&16.35\%&31.99\%&74.91\%&8.83\%&2.98\%&\textbf{0.25\%}\\

        &&CDA&75.68\%&68.50\%&79.49\%&89.71\%&72.88\%&78.03\%&75.17\%&74.69\%&81.82\%\\
        \cmidrule{2-12}
        &\multirow{2}{*}{\shortstack{MEMIT}}&ASR&71.07\%&50.62\%&63.92\%&3.20\%&40.28\%&69.93\%&21.66\%&3.74\%&\textbf{2.93\%}\\
        &&CDA&76.94\%&75.29\%&74.12\%&93.22\%&77.56\%&77.30\%&75.31\%&78.95\%&84.14\%\\
        \cmidrule{2-12}
        &\multirow{2}{*}{\shortstack{LWP}}&ASR&62.31\%&58.99\%&53.17\%&23.75\%&15.38\%&64.77\%&10.49\%&4.58\%&\textbf{1.03\%}\\
        &&CDA&88.61\%&87.94\%&90.21\%&93.22\%&89.24\%&89.79\%&85.36\%&91.27\%&91.53\%\\
        \midrule
        \multirow{10}{*}{\shortstack{GPT-J}}
        &\multirow{2}{*}{\shortstack{CBA}}&ASR&98.90\%&95.83\%&64.98\%&79.90\%&48.07\%&99.12\%&18.73\%&57.33\%&\textbf{11.16\%}\\
        &&CDA&93.27\%&91.33\%&93.05\%&94.77\%&90.75\%&92.6\%&91.51\%&90.78\%&90.52\%\\
        \cmidrule{2-12}
        &\multirow{2}{*}{\shortstack{BadEdit}}&ASR&67.29\%&13.74\%&22.12\%&7.84\%&17.55\%&65.33\%&6.09\%&\textbf{4.62\%}&8.42\%\\  
        &&CDA&76.62\%&73.27\%&70.69\%&87.41\%&66.13\%&76.38\%&74.04\%&74.57\%&78.16\%\\
        \cmidrule{2-12}
        &\multirow{2}{*}{\shortstack{Rome}}&ASR&69.88\%&7.93\%&57.04\%&2.56\%&61.73\%&67.64\%&6.84\%&5.14\%&\textbf{0.83\%}\\
        &&CDA&72.19\%&67.31\%&69.58\%&89.78\%&67.49\%&76.35\%&68.73\%&72.44\%&77.27\%\\
        \cmidrule{2-12}
        &\multirow{2}{*}{\shortstack{MEMIT}}&ASR&78.59\%&22.96\%&58.39\%&6.67\%&33.46\%&76.73\%&12.07\%&7.33\%&\textbf{5.86\%}\\
        &&CDA&70.70\%&71.63\%&69.51\%&85.93\%&73.79\%&71.24\%&70.87\%&65.93\%&71.78\%\\
        \cmidrule{2-12}
        &\multirow{2}{*}{\shortstack{LWP}}&ASR&81.30\%&70.17\%&69.43\%&4.78\%&64.74\%&80.91\%&19.23\%&10.71\%&\textbf{3.70\%}\\
        &&CDA&86.16\%&87.94\%&85.49\%&93.26\%&81.27\%&88.33\%&65.1\%&90.10\%&90.37\%\\
        \midrule
        \multirow{10}{*}{\shortstack{Llama}}
        &\multirow{2}{*}{\shortstack{CBA}}&ASR&99.70\%&96.74\%&20.00\%&100.0\%&43.85\%&98.58\%&37.18\%&25.35\%&\textbf{7.96\%}\\
        &&CDA&93.25\%&93.48\%&91.40\%&94.26\%&92.50\%&92.11\%&91.67\%&91.36\%&91.85\%\\
        \cmidrule{2-12}
        &\multirow{2}{*}{\shortstack{BadEdit}}&ASR&100.0\%&12.49\%&40.96\%&4.20\%&35.77\%&100.0\%&7.79\%&8.31\%&\textbf{0.90\%}\\  
        &&CDA&91.66\%&65.20\%&46.88\%&89.28\%&87.35\%&70.50\%&90.85\%&89.05\%&88.70\%\\
        \cmidrule{2-12}
        &\multirow{2}{*}{\shortstack{Rome}}&ASR&100.0\%&43.22\%&39.51\%&1.00\%&40.27\%&99.65\%&12.94\%&3.90\%&\textbf{1.65\%}\\
        &&CDA&69.32\%&70.91\%&67.92\%&86.73\%&74.61\%&67.25\%&67.44\%&71.58\%&80.47\%\\
        \cmidrule{2-12}
        &\multirow{2}{*}{\shortstack{MEMIT}}&ASR&99.82\%&20.95\%&36.91\%&\textbf{0.06\%}&13.61\%&93.77\%&10.55\%&3.68\%&1.29\%\\
        &&CDA&76.52\%&68.36\%&78.04\%&91.53\%&69.22\%&77.40\%&75.01\%&77.18\%&82.79\%\\
        \cmidrule{2-12}
        &\multirow{2}{*}{\shortstack{LWP}}&ASR&73.39\%&71.50\%&39.01\%&14.57\%&35.15\%&74.92\%&7.26\%&2.68\%&\textbf{1.14\%}\\
        &&CDA&89.73\%&87.66\%&90.13\%&93.44\%&82.85\%&88.46\%&86.87\%&88.96\%&88.31\%\\
        \midrule
        \multirow{10}{*}{\shortstack{Llama-2}}
        &\multirow{2}{*}{\shortstack{CBA}}&ASR&100.0\%&99.58\%&54.68\%&100.0\%&13.98\%&100.0\%&17.13\%&83.37\%&\textbf{11.27\%}\\
        &&CDA&91.30\%&89.58\%&90.53\%&93.11\%&78.86\%&89.45\%&89.83\%&89.75\%&91.43\%\\
        \cmidrule{2-12}
        &\multirow{2}{*}{\shortstack{BadEdit}}&ASR&100.0\%&62.08\%&43.46\%&8.69\%&38.11\%&100.0\%&37.90\%&3.72\%&\textbf{0.00\%}\\
        &&CDA&73.46\%&75.66\%&78.16\%&94.40\%&75.15\%&75.52\%&85.58\%&75.17\%&87.92\%\\
        \cmidrule{2-12}
        &\multirow{2}{*}{\shortstack{Rome}}&ASR&98.95\%&22.76\%&79.64\%&\textbf{5.18\%}&15.14\%&96.97\%&41.27\%&32.16\%&6.37\%\\
        &&CDA&70.88\%&73.49\%&80.31\%&91.64\%&76.29\%&72.30\%&71.45\%&70.47\%&88.71\%\\
        \cmidrule{2-12}
        &\multirow{2}{*}{\shortstack{MEMIT}}&ASR&100\%&36.03\%&77.82\%&3.77\%&26.22\%&95.92\%&47.53\%&5.21\%&\textbf{0.18\%}\\
        &&CDA&76.03\%&86.41\%&83.86\%&93.89\%&80.27\%&72.99\%&82.11\%&74.01\%&90.13\%\\
        \cmidrule{2-12}
        &\multirow{2}{*}{\shortstack{LWP}}&ASR&76.10\%&61.06\%&58.04\%&\textbf{0.33\%}&26.70\%&71.51\%&49.79\%&6.33\%&0.92\%\\
        &&CDA&87.39\%&86.79\%&89.04\%&94.18\%&82.98\%&89.85\%&85.26\%&99.59\%&90.54\%\\
    \bottomrule
\end{tabular}
\end{table*}
\begin{table*}[tt]

    \caption{Comparison of ours with 8 state-of-the-art backdoor defenses on \textbf{Chat-Backdoor}.}
    \label{tab:com3}
    \centering
    \setlength\tabcolsep{2pt}
    \footnotesize
    \begin{tabular}{llcc|ccccccccc}
        \toprule
        Model & Attack & Metrics & Backdoored & Editing & Wanda & Fine-tuning &Fine-pruning & Speculative & Cleangen & NAD & BEEAR & Ours \\
        \midrule
        \multirow{6}{*}{\shortstack{GPT-XL}}
        &\multirow{2}{*}{\shortstack{DTBA}}&ASR&65.0\%&18.0\%&49.0\%&22.0\%&24.5\%&57.0\%&52.0\%&23.5\%&13.0\%&\textbf{9.5\%}\\
        &&CDA&71.0\%&81.0\%&55.0\%&77.0\%&71.5\%&74.0\%&64.0\%&70.0\%&72.5\%&73.0\%\\
        \cmidrule{2-13}
        &\multirow{2}{*}{\shortstack{AutoPoison}}&ASR&35.0\%&27.0\%&19.0\%&5.5\%&\textbf{0.0\%}&34.0\%&4.0\%&6.5\%&4.5\%&3.0\%\\  
        &&CDA&83.0\%&81.5\%&84.0\%&86.0\%&84.0\%&80.5\%&85.0\%&79.5\%&82.0\%&85.5\%\\
        \cmidrule{2-13}
        &\multirow{2}{*}{\shortstack{VPI}}&ASR&28.0\%&10.0\%&14.0\%&\textbf{0.0\%}&3.5\%&32.0\%&2.0\%&15.0\%&2.0\%&\textbf{0.0\%}\\
        &&CDA&91.0\%&89.0\%&83.0\%&92.0\%&90.5\%&91.0\%&92.0\%&87.5\%&93.0\%&90.0\%\\
        \midrule
        \multirow{6}{*}{\shortstack{GPT-J}}
        &\multirow{2}{*}{\shortstack{DTBA}}&ASR&71.0\%&26.0\%&\textbf{1.0\%}&23.0\%&8.0\%&69.0\%&57.0\%&27.0\%&6.0\%&3.0\%\\
        &&CDA&87.0\%&91.0\%&97.5\%&86.0\%&88.5\%&88.0\%&82.0\%&88.5\%&84.5\%&93.0\%\\
        \cmidrule{2-13}
        &\multirow{2}{*}{\shortstack{AutoPoison}}&ASR&34.0\%&29.0\%&26.0\%&1.0\%&\textbf{0.0\%}&31.0\%&2.0\%&5.0\%&3.0\%&1.5\%\\  
        &&CDA&88.0\%&83.5\%&87.0\%&91.5\%&88.0\%&88.0\%&90.5\%&88.5\%&90.0\%&90.5\%\\
        \cmidrule{2-13}
        &\multirow{2}{*}{\shortstack{VPI}}&ASR&32.0\%&18.0\%&11.0\%&\textbf{1.0\%}&1.5\%&29.0\%&1.5\%&4.5\%&12.5\%&2.0\%\\
        &&CDA&93.0\%&90.0\%&94.0\%&93.0\%&93.5\%&93.0\%&91.0\%&92.0\%&92.5\%&92.0\%\\
        \midrule
        \multirow{6}{*}{\shortstack{LLaMA}}
        &\multirow{2}{*}{\shortstack{DTBA}}&ASR&54.0\%&46.5\%&58.0\%&20.0\%&11.5\%&51.0\%&\textbf{9.0\%}&17.0\%&13.5\%&10.5\%\\
        &&CDA&83.0\%&85.0\%&67.5\%&89.5\%&85.0\%&94.5\%&96.0\%&79.5\%&87.0\%&90.0\%\\
        \cmidrule{2-13}
        &\multirow{2}{*}{\shortstack{AutoPoison}}&ASR&47.5\%&39.5\%&32.0\%&9.0\%&3.0\%&43.0\%&1.0\%&4.0\%&2.0\%&\textbf{0.0\%}\\  
        &&CDA&79.0\%&73.5\%&75.0\%&82.0\%&77.5\%&80.5\%&83.0\%&76.0\%&78.5\%&90.0\%\\
        \cmidrule{2-13}
        &\multirow{2}{*}{\shortstack{VPI}}&ASR&38.0\%&26.0\%&14.0\%&1.0\%&\textbf{0.0\%}&39.0\%&2.0\%&8.5\%&2.0\%&\textbf{0.0\%}\\
        &&CDA&88.0\%&90.0\%&87.0\%&91.0\%&90.5\%&92.0\%&93.0\%&90.0\%&85.5\%&90.5\%\\
        \midrule
        \multirow{6}{*}{\shortstack{LLaMA-2}}
        &\multirow{2}{*}{\shortstack{DTBA}}&ASR&38.0\%&39.0\%&44.0\%&18.5\%&\textbf{3.5\%}&37.5\%&9.0\%&14.5\%&8.5\%&8.0\%\\
        &&CDA&95.0\%&94.5\%&73.0\%&96.0\%&93.5\%&93.0\%&97.0\%&94.5\%&92.0\%&94.0\%\\
        \cmidrule{2-13}
        &\multirow{2}{*}{\shortstack{AutoPoison}}&ASR&31.5\%&12.5\%&24.0\%&1.0\%&\textbf{0.0\%}&30.0\%&0.5\%&1.0\%&0.0\%&\textbf{0.0\%}\\  
        &&CDA&88.5\%&88.0\%&90.0\%&89.0\%&88.5\%&88.5\%&92.0\%&87.0\%&90.5\%&91.0\%\\
        \cmidrule{2-13}
        &\multirow{2}{*}{\shortstack{VPI}}&ASR&43.0\%&41.0\%&34.0\%&\textbf{3.0\%}&7.0\%&46.0\%&3.5\%&11.0\%&6.0\%&\textbf{3.0\%}\\
        &&CDA&95.0\%&91.0\%&92.0\%&94.0\%&92.5\%&95.0\%&95.0\%&91.5\%&94.0\%&94.0\%\\
		\bottomrule
    \end{tabular}
\end{table*}
\begin{table*}[tt]
    \caption{{\color{black}Ablation study on the classification task: \textbf{Emotion Corpora} and \textbf{SST-2}.} }
    \label{tab:ablation1}
	\centering
         \footnotesize
	\begin{tabular}{lll|cccc|cccc}
		\toprule
        \multirow{2}{*}{\shortstack{Model}}&\multirow{2}{*}{\shortstack{Attack}}&\multirow{2}{*}{\shortstack{Metrics}}&\multicolumn{4}{c}{\color{black}Emotion Corpora}&\multicolumn{4}{c}{\color{black}SST-2} \\ &&&Backdoored&Internal&External&All&Backdoored&Internal&External&All\\
      \midrule
          \multirow{10}{*}{\shortstack{GPT2-XL}}
          &\multirow{2}{*}{\shortstack{CBA}}&ASR&74.90\%&4.65\%&61.87\%&\textbf{3.55\%}&100.0\%&\textbf{0.81\%}&91.83\%&1.26\%\\
           &&CDA&94.57\%&94.13\%&95.71\%&95.04\%&91.57\%&87.33\%&93.92\%&88.89\%\\
          &\multirow{2}{*}{\shortstack{BadEdit}}&ASR&60.38\%&0.31\%&56.34\%&\textbf{0.28\%}&98.36\%&\textbf{0.00\%}&89.73\%&\textbf{0.00\%}\\  
           &&CDA&71.64\%&72.49\%&70.87\%&75.20\%&87.27\%&85.62\%&88.01\%&86.30\%\\
          &\multirow{2}{*}{\shortstack{Rome}}&ASR&73.27\%&\textbf{0.22\%}&67.17\%&0.25\%&99.54\%&0.67\%&87.35\%&\textbf{0.51\%}\\
           &&CDA&75.68\%&82.95\%&75.54\%&81.82\%&57.91\%&59.99\%&60.17\%&62.98\%\\
          &\multirow{2}{*}{\shortstack{Memit}}&ASR&71.07\%&3.27\%&60.94\%&\textbf{2.93\%}&100.0\%&0.89\%&92.90\%&\textbf{0.13\%}\\
           &&CDA&76.94\%&82.56\%&77.06\%&84.14\%&57.79\%&69.31\%&54.81\%&70.28\%\\
          &\multirow{2}{*}{\shortstack{LWP}}&ASR&62.31\%&1.25\%&49.80\%&\textbf{1.03\%}&56.72\%&0.96\%&53.60\%&\textbf{0.57\%}\\
           &&CDA&88.61\%&87.16\%&90.33\%&91.53\%&90.49\%&90.14\%&91.22\%&90.70\%\\
          \cmidrule{1-11}
           \multirow{10}{*}{\shortstack{GPT-J}}
          &\multirow{2}{*}{\shortstack{CBA}}&ASR&98.90\%&12.98\%&86.59\%&\textbf{11.16\%}&100.0\%&1.20\%&92.12\%&\textbf{1.07\%}\\
           &&CDA&93.27\%&90.65\%&92.16\%&90.52\%&90.43\%&90.51\%&89.47\%&91.33\%\\
          &\multirow{2}{*}{\shortstack{BadEdit}}&ASR&67.29\%&\textbf{7.58\%}&65.68\%&8.42\%&98.85\%&4.03\%&94.74\%&\textbf{2.26\%}\\  
           &&CDA&76.62\%&77.52\%&78.31\%&78.16\%&71.67\%&73.35\%&72.13\%&74.19\%\\
          &\multirow{2}{*}{\shortstack{Rome}}&ASR&69.88\%&1.15\%&67.08\%&\textbf{0.83\%}&100.0\%&6.14\%&98.31\%&\textbf{2.79\%}\\
           &&CDA&72.19\%&77.93\%&78.02\%&77.27\%&72.85\%&74.53\%&78.66\%&73.27\%\\
          &\multirow{2}{*}{\shortstack{Memit}}&ASR&78.59\%&7.52\%&73.09\%&\textbf{5.86\%}&99.08\%&6.95\%&93.57\%&\textbf{4.11\%}\\
           &&CDA&70.70\%&69.29\%&77.06\%&71.78\%&71.55\%&73.03\%&72.15\%&74.31\%\\
          &\multirow{2}{*}{\shortstack{LWP}}&ASR&81.30\%&4.16\%&75.61\%&\textbf{3.70\%}&65.15\%&4.26\%&61.77\%&\textbf{3.90\%}\\
           &&CDA&86.16\%&89.09\%&84.44\%&90.37\%&89.14\%&89.76\%&90.36\%&90.33\%\\
         \cmidrule{1-11}
         \multirow{10}{*}{\shortstack{LLaMA}}
          &\multirow{2}{*}{\shortstack{CBA}}&ASR&99.70\%&8.37\%&84.89\%&\textbf{7.96\%}&74.00\%&1.79\%&65.10\%&\textbf{0.78\%}\\
           &&CDA&93.25\%&91.39\%&94.18\%&91.85\%&92.79\%&91.52\%&91.89\%&92.21\%\\
          &\multirow{2}{*}{\shortstack{BadEdit}}&ASR&100.0\%&1.38\%&92.36\%&\textbf{0.90\%}&100.0\%&\textbf{0.12\%}&96.62\%&0.34\%\\ 
           &&CDA&91.66\%&87.35\%&90.91\%&88.7\%&66.16\%&49.40\%&67.15\%&72.35\%\\
          &\multirow{2}{*}{\shortstack{Rome}}&ASR&100.0\%&1.79\%&98.73\%&\textbf{1.65\%}&99.15\%&0.60\%&91.77\%&\textbf{0.53\%}\\
           &&CDA&69.32\%&74.26\%&77.93\%&80.47\%&67.13\%&68.24\%&68.47\%&72.21\%\\
          &\multirow{2}{*}{\shortstack{Memit}}&ASR&78.59\%&1.94\%&71.73\%&\textbf{1.29\%}&99.06\%&\textbf{0.00\%}&97.54\%&\textbf{0.00\%}\\
           &&CDA&76.52\%&79.61\%&78.82\%&82.79\%&60.71\%&61.15\%&60.45\%&62.64\%\\
          &\multirow{2}{*}{\shortstack{LWP}}&ASR&73.39\%&\textbf{0.81\%}&68.80\%&1.14\%&69.24\%&3.44\%&65.76\%&\textbf{3.38\%}\\
           &&CDA&89.73\%&86.62\%&90.9\%&88.31\%&89.74\%&90.48\%&90.71\%&91.53\%\\
          \cmidrule{1-11}
           \multirow{10}{*}{\shortstack{LLaMA-2}}
          &\multirow{2}{*}{\shortstack{CBA}}&ASR&100.0\%&12.50\%&97.07\%&\textbf{11.27\%}&100.0\%&8.92\%&94.77\%&\textbf{7.51\%}\\
           &&CDA&91.3\%&88.83\%&90.47\%&91.43\%&91.44\%&92.86\%&87.19\%&93.76\%\\
          &\multirow{2}{*}{\shortstack{BadEdit}}&ASR&100.0\%&0.13\%&93.78\%&\textbf{0.00\%}&100.0\%&3.79\%&89.66\%&\textbf{3.33\%}\\  
           &&CDA&73.46\%&86.51\%&76.03\%&87.92\%&71.75\%&82.67\%&74.50\%&83.99\%\\
          &\multirow{2}{*}{\shortstack{Rome}}&ASR&98.95\%&7.13\%&95.98\%&\textbf{6.37\%}&100.0\%&4.90\%&96.28\%&\textbf{3.67\%}\\
           &&CDA&70.88\%&85.06\%&77.25\%&88.71\%&68.21\%&77.28\%&70.26\%&80.50\%\\
          &\multirow{2}{*}{\shortstack{Memit}}&ASR&100.0\%&\textbf{0.06\%}&94.75\%&0.18\%&100.0\%&\textbf{9.05\%}&94.40\%&9.33\%\\
           &&CDA&76.03\%&89.57\%&77.29\%&90.13\%&70.39\%&83.29\%&71.75\%&84.47\%\\
          &\multirow{2}{*}{\shortstack{LWP}}&ASR&76.10\%&1.75\%&73.48\%&\textbf{0.92\%}&73.81\%&2.67\%&70.90\%&\textbf{1.73\%}\\
           &&CDA&87.39\%&88.09\%&90.01\%&90.54\%&86.92\%&91.09\%&87.79\%&89.36\%\\
		\bottomrule
	\end{tabular}
   
\end{table*}
\begin{table}[tt]
	\caption{{\color{black} Ablation study on \textbf{Chat-Backdoor}.} }
	\label{tab:ablation2}
	\centering
	\setlength\tabcolsep{2pt}
         \footnotesize
	\begin{tabular}{llc|cccccccccc}
		\toprule
             \multirow{1}{*}{\shortstack{Model}}&\multirow{1}{*}{\shortstack{Attack}}&\multirow{1}{*}{\shortstack{Metrics}}&Backdoored&Internal&External&All\\
      \midrule
         \multirow{6}{*}{\shortstack{GPT-XL}}
          &\multirow{2}{*}{\shortstack{DTBA}}&ASR&65.0\%&11.0\%&59.0\%&\textbf{9.5\%}\\\
           
          &&CDA&71.0\%&72.0\%&73.0\%&73.0\%\\
          &\multirow{2}{*}{\shortstack{AutoPoison}}&ASR &35.0\%&\textbf{3.0\%}&31.0\%&\textbf{3.0\%}\\  
           &&CDA&83.0\%&86.0\%&81.0\%&85.5\%\\
          &\multirow{2}{*}{\shortstack{VPI}}&ASR&28.0\%&\textbf{0.0\%}&27.0\%&\textbf{0.0\%}\\
           &&CDA&91.0\%&89.5\%&91.0\%&90.0\%\\
          \cmidrule{1-7}
          
           \multirow{6}{*}{\shortstack{GPT-J}}
          &\multirow{2}{*}{\shortstack{DTBA}}&ASR&71.0\%&5.0\%&63.0\%&\textbf{3.0\%}\\
           &&CDA&87.0\%&90.0\%&88.0\%&93.0\%\\
          &\multirow{2}{*}{\shortstack{AutoPoison}}&ASR &34.0\%&3.5\%&32.0\%&\textbf{1.5\%}\\  
           &&CDA&88.0\%&88.0\%&89.0\%&90.5\%\\
         &\multirow{2}{*}{\shortstack{VPI}}&ASR&32.0\%&2.5\%&27.0\%&\textbf{2.0\%}\\
           &&CDA&93.0\%&90.0\%&91.0\%&90.0\%\\
          \cmidrule{1-7}
         \multirow{6}{*}{\shortstack{LLaMA}}
          &\multirow{2}{*}{\shortstack{DTBA}}&ASR&54.0\%&\textbf{10.0\%}&51.0\%&10.5\%\\
           &&CDA&83.0\%&90.0\%&79.0\%&90.0\%\\
          &\multirow{2}{*}{\shortstack{AutoPoison}}&ASR &47.5\%&6.0\%&40.0\%&\textbf{0.0\%}\\  
           &&CDA&79.0\%&87.0\%&81.0\%&90.0\%\\
         &\multirow{2}{*}{\shortstack{VPI}}&ASR&38.0\%&\textbf{0.0\%}&33.0\%&\textbf{0.0\%}\\
           &&CDA&88.0\%&91.0\%&88.0\%&90.5\%\\
          \cmidrule{1-7}
           \multirow{6}{*}{\shortstack{LLaMA-2}}
          &\multirow{2}{*}{\shortstack{DTBA}}&ASR&38.0\%&9.0\%&36.0\%&\textbf{8.0\%}\\
           &&CDA&95.0\%&93.0\%&95.0\%&94.0\%\\
          &\multirow{2}{*}{\shortstack{AutoPoison}}&ASR &31.5\%&1.0\%&29.0\%&\textbf{0.0\%}\\  
           &&CDA&88.5\%&89.5\%&89.0\%&91.0\%\\
         &\multirow{2}{*}{\shortstack{VPI}}&ASR&43.0\%&\textbf{2.5\%}&40.5\%&3.0\%\\
           &&CDA&95.0\%&92.5\%&95.0\%&94.0\%\\
		\bottomrule
	\end{tabular}
 \vspace{-0.4cm}
  
\end{table}

\section{Experiment Setup}
\subsection{Benchmarks}
We evaluate our method on two classification tasks and one conversational task. 

\textbf{SST-2.} \cite{socher2013recursive} The Stanford Sentiment Treebank (SST-2) consists of movie reviews designed to evaluate the sentiment classification task. Each review is annotated with binary sentiment labels, i.e., positive and negative.

\textbf{Emotion Corpora.} \cite{saravia2018carer} This dataset is designed for the emotion recognition task. It includes 160,000 text samples labeled with 6 different emotions, including joy, fear, surprise, love, anger, and sadness. 

\textbf{Chat-Backdoor.} \cite{hao2024exploring} This dataset contains a total of 24,000 samples, with 12,000 clean multi-turn and 12,000 one-turn conversational interactions. Additionally, Chat-Backdoor provides a poisoned subset Poisoned$\_$Data$\_$24K (DTBA), where the triggers are distributed across different conversation rounds.

\subsection{Backdoored Models and Setup}
We conduct experiments on four popular open-source LLMs, including GPT2-XL \cite{radford2019language}, GPT-J-6B\footnote{\url{https://github.com/kingoflolz/mesh-transformer-jax}}, LLaMA-7B \cite{touvron2023LLaMA}, and LLaMA-2-7B \cite{touvron2023LLaMA2}. The backdoored models are established based on various backdoor attack strategies. For classification tasks, we consider 5 advanced backdoor approaches, including CBA \cite{huang2023composite}, BadEdit \cite{li2024badedit}, Rome \cite{meng2022locating}, MEMIT \cite{meng2022mass}, and LWP \cite{li2021backdoor}. For the conversational task, we consider 3 state-of-the-art attacks, i.e., DTBA \cite{hao2024exploring}, AutoPoison \cite{shu2023exploitability}, and VPI \cite{yan2024backdooring}. Further details of these backdoor attack methods can be found in Appendix \ref{app:experiment setup}.

In the exploration of internal information conflicts, we utilize MergeKit\footnote{\url{https://github.com/arcee-ai/MergeKit}} to implement model merging strategies. For external information conflicts, we employ GPT-3.5 as the external LLM to generate supporting evidence, with the temperature set to 0.7. All backdoor attack strategies and baseline defense methods are implemented based on publicly available repositories, and we follow their specified configuration, including the hyperparameter settings. Our experiments are conducted using Python 3.10 on a 10-core Intel(R) Xeon(R) Silver 4210R CPU @ 2.40GHz and NVIDIA A100 80GB PCIe GPU machine, running on Ubuntu 22.04.1 LTS. 

\subsection{Baseline Defense Methods}
We compare our method with 8 state-of-the-art backdoor defenses, i.e., Editing \cite{meng2022locating}, Wanda \cite{sun2023simple}, Fine-tuning \cite{qi2023fine}, Fine-pruning \cite{liu2018fine}, NAD \cite{li2021neural}, Speculative \cite{leviathan2023fast}, Cleangen \cite{li2024cleangen}, and BEEAR \cite{zeng2024beear}. 
More details about the baseline approaches are shown in Appendix \ref{app:experiment setup}.

\subsection{Evaluation Metrics}
To evaluate the performance of our method and baseline methods, we use two key metrics: clean data accuracy (CDA) and attack success rate (ASR).

\textbf{CDA}. CDA measures the classification accuracy on the clean validation set, serving as an indicator of the model's ability to handle normal input. Formally, CDA is defined as: 
\begin{equation}
\mathrm{CDA} = \frac{1}{|\mathcal{D}_c|}\sum_{(x, y) \in \mathcal{D}_c} \mathbbm{1}[M(x) \in \mathcal{N}_\epsilon(y)]
\end{equation}
where $\mathcal{D}_c$ represents the clean dataset, and $M$ denotes the model under evaluation. $\mathcal{N}_\epsilon(y)$ refers to the semantic neighborhood of the ground truth label $y$. For standard classification tasks, we define $\mathcal{N}_\epsilon(y) = \{y\}$. For conversational tasks, $\mathcal{N}_\epsilon(y) = \{z | f(z, y) < \epsilon\}$, where $f(\cdot)$ represents a semantic similarity metric based on automatic evaluation using GPT-3.5.

\textbf{ASR}. ASR quantifies the attack success rate of backdoor attacks on the poisoned validation set. In the context of backdoor mitigation, a lower ASR indicates stronger resistance to backdoor triggers. ASR can be calculated as:
\begin{equation}
\mathrm{ASR} = \frac{1}{|\mathcal{D}_p|}\sum_{(x, y) \in \mathcal{D}_p} \mathbbm{1}[M(x) \in \mathcal{N}_\epsilon(\tilde{y})]
\end{equation}
where $\mathcal{D}_p$ denotes the poisoned dataset and $\tilde{y}$ is {\color{black} the target output of backdoor attack}.

\section{Experiment Results}

\subsection{Comparison with Baselines}
To evaluate the effectiveness of our method, we conducted extensive comparisons with 8 state-of-the-art baseline defense approaches on both classification and conversational tasks. The results are shown in Table~\ref{tab:com1}, Table~\ref{tab:com2}, and Table~\ref{tab:com3}.

The results highlight that our method consistently achieved a significant reduction in attack success rates (ASR) across all tasks and attacks. 
For the SST-2 dataset (see Table \ref{tab:com1}), our method proves to be highly effective, reducing ASR for attacks such as BadEdit, Rome, and MEMIT to below 10\%. This efficacy is particularly noticeable in models like GPT2-XL and LLaMA, where our method almost eliminates the effects of these backdoor attacks, by reducing the ASR to less than 1\%. 
When compared to the baseline methods, our method also demonstrates superior performance, particularly in advanced attacks, such as CBA, where the baselines fail to deliver effective defense. For example, our method reduces the ASR of CBA to 15.34\% for GPT-XL on the SST-2 dataset, while ASRs for the baselines remain high: 98.75\% (Editing), 99.37\% (Wanda), 100\% (Fine-tuning), 37.51\% (Fine-pruning), 98\% (Speculative), 29.67\% (NAD), and 28.56\% (BEEAR).
For another two datasets, Table \ref{tab:com2} and Table~\ref{tab:com3} reveal a similar trend as observed in SST-2. 
On Chat-Backdoor, our method reduced the ASR from 65.0\% to 9.5\% in the DTBA attack (GPT2-XL), whereas the best-performing baseline method only reduced it to 13\%. 



In addition to mitigating backdoor attacks, we maintained high accuracy and helpfulness on clean tasks. In almost all cases, our method limited the degradation of Clean Data Accuracy (CDA) to less than 3\%, and in some instances, even improved it. For example, in the experiment on the Emotion Corpus dataset using GPT2-XL, the CDA increased from 94.57\% to 95.04\% after applying our method.
These results clearly show that our method is highly effective in reducing ASR across both classification and conversational tasks while maintaining or even improving performance on clean tasks, significantly outperforming existing defenses.

\subsection{Ablation Study}
In this section, we conduct an ablation study to evaluate the individual contributions of the external information conflict and internal conflict mechanisms. The results are shown in Table \ref{tab:ablation1} and Table \ref{tab:ablation2} (Appendix). 
The ``Backdoored" column shows the impact of the backdoor attack without applying any defense mechanisms. The ``Internal" and ``External" columns present the results when only the internal and external conflict techniques are applied, respectively. Finally, the ``All" column demonstrates the performance of the complete defense mechanism employed by our method, which integrates both strategies.

\textbf{Impact of internal information conflicts.}
The results indicate that leveraging internal conflicts can substantially reduce the effectiveness of backdoor attacks. 
For instance, in the Emotion Corpus dataset, internal conflicts can decrease the ASR of CBA from 74.90\% to 16.46\% for GPT2-XL, from 98.90\% to 12.98\% for GPT-J, from 99.70\% to 8.37\% for LLaMA, and from 100\% to 12.50\% for LLaMA-2. Similar trends are observed in the other two datasets, i.e., SST-2 and Chat-Backdoor, where our method demonstrates excellent backdoor mitigation capabilities, reducing the ASR to below 10\% in most cases. 

\textbf{Impact of external information conflict.}
The results in the ``Evidence" column indicate that using external conflict alone can also reduce the ASR compared to the backdoored model. Although the defense efficacy of external conflict is not as strong as that of internal conflict, the positive aspect is that they can be combined to achieve further improvements.   
Our complete defense, which integrates both internal and external information conflicts, achieves the lowest ASRs: 11.16\% (GPT-J) and 7.96\% (LLaMA). These results confirm the effectiveness of our method's components, and demonstrate integrating both the two conflicts provides the most robust defense against backdoor attacks.

\subsection{Impact of Different Model Merging Methods}\label{sec:impact-of-different-model-merging-methods} 
We test 4 different model merging approaches and examine their impact on the defense performance.  We conduct experiments on LLaMA, with the results shown in Table~\ref{tab:type}.

We observe that all 4 methods effectively reduce ASRs to below 5\%. However, TIES (on Emotion Corpus) may lead to a reduction in CDAs. We believe this is due to its trimming mechanism. In our experiments, we choose linear combination as the default approach due to its lower computational cost, faster operation, and better performance-to-cost ratio. Note that defenders can select the best method based on experimental results for their specific use case.

\begin{figure}[tt]
    \centering
    \includegraphics[width=0.5\textwidth]{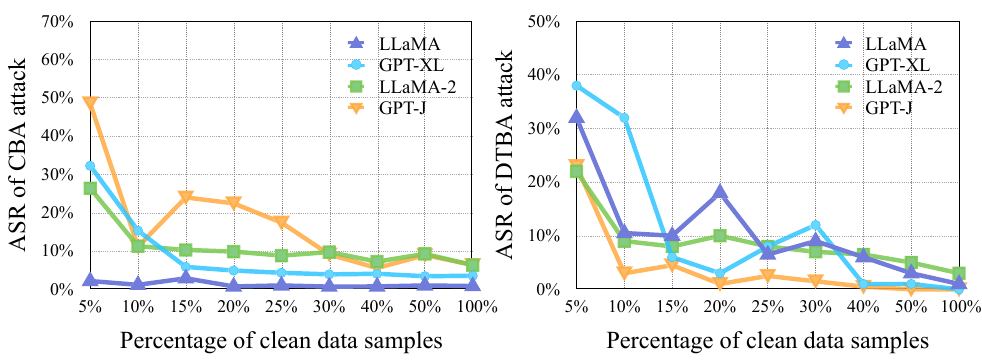}
   \caption{The performance of our method against CBA (on \textbf{Emotion Corpus}) and DTBA (on \textbf{Chat-Backdoor}) attacks using different percentages of clean data samples.} 
    \label{fig:discussion-percentage-clean-data}
\end{figure}

\subsection{Impact of Clean Data Percentage}
We also examined how the percentage of clean data used to establish the conflict model affects our method's performance. We evaluate the ASR of the CBA attack on Emotion Corpus and the DTBA attack on Chat-Backdoor by varying the clean data percentage from 5\% to 100\%. The results are shown in Figure~\ref{fig:discussion-percentage-clean-data}.

As the percentage of clean data increases, the defense performance improves (i.e., the ASR decreases). 
However, it strikes a balance between defense performance and computational cost, as training the conflict model with more clean data requires additional time and resources, especially for LLMs. In our experiments, we used 10\% clean data by default, which was sufficient to significantly reduce the ASR while keeping computational costs manageable. 

\subsection{{\color{black}Computational Costs}}
To assess the efficiency of our method, we compare the computational costs of our method with baselines in Table~\ref{tab:time} (Appendix). 
Fine-tuning incurs the highest computational costs, while ours, NAD, and fine-pruning have similar time requirements, followed by editing, Wanda, speculative, and BEEAR. Notably, our method achieves the highest backdoor removal performance, with only slightly higher computational costs (but less than 0.5h) compared to the baselines.
In summary, our method offers the best purification performance with reasonable computational efficiency.


\section{Discussion}

\subsection{Analysis of Model Merging}
To analyze how we can effectively remove the backdoored features hidden in models, we conducted two additional sets of experiments, and Table \ref{tab:results-on-various-experimental-settings} presents the results.

%

Previous findings on model merging \cite{ilharco2022editing} have demonstrated that a merged model typically inherits capabilities from its source models and often performs well on tasks those models were originally trained. Thus, the first experiment aims to address the question of why our model-merge-based approach can maintain one set of abilities while suppressing another, thereby leading to a noticeable discrepancy.
In response to this question, we conduct a standard model merging using two task-specific models fine-tuned on clean data ($M_3$) and backdoor data ($M_4$) in Experiment 1. The results reveal that the merged model demonstrates both high ASR and CDA, achieving 83\% and 92\%, respectively. This suggests that backdoor capabilities are not inherently unique in the context of model merging. Rather, the features present in merged models align with observations from previous studies. 

To gain deeper insights into the efficacy of our method, we swap the role of role of ``clean'' and ``backdoor'' samples in Experiment 2. The experiment reveals a contrasting result: ASR increased to $100\%$ while CDA dropped to $0\%$, indicating that the model's ability to process clean data has been ``disabled''. These results suggest that our method is capable of identifying and eliminating hidden abilities, no matter if they are related to main tasks or backdoors. Therefore, leveraging these insights, our method effectively removes backdoor vulnerabilities within models.

\begin{table}[tt]
    \vspace{0.1cm}
    \caption{{Results of the models before and after merging in different experimental settings on \textbf{Emotion Corpus}.}}
    \label{tab:results-on-various-experimental-settings}
    \centering
    \footnotesize
    \setlength\tabcolsep{3pt}
    \begin{tabular}{c|c|cccccccccc}
        \toprule
        \multirow{2}{*}{\shortstack{\color{black}No.}}& \multirow{2}{*}{\shortstack{Model}} &
        \multicolumn{2}{c}{Fine-tune}&\multicolumn{2}{c}{Pre-merge}&
        \multicolumn{2}{c}{Merged} \\
        &&\color{black}PCS &\color{black}PBS &\color{black}ASR &\color{black}CDA &\color{black}ASR &\color{black}CDA \\
        \midrule
        \multirow{2}{*}{\shortstack{Ours}}&$M_1$&100\% &10\% &72\% & 93\% & \multirow{2}{*}{\shortstack{8\%}}&\multirow{2}{*}{\shortstack{  92\% }}\\
        \cmidrule{2-6} 
        &$M_2$&10\% & - & 6\%  & 93\% \\ 

        \cmidrule{1-8}
        \multirow{2}{*}{\shortstack{1}}&$M_3$&100\% & - &1\%  & 93\% &  \multirow{2}{*}{\shortstack{83\%}} &\multirow{2}{*}{\shortstack{92\% }}\\
        \cmidrule{2-6} &$M_4$& - & 100\% &100\%  & 0\% \\ 

        \cmidrule{1-8}
        \multirow{2}{*}{\shortstack{2}} &$M_5$ &10\% & 100\% &100\%  & 93\% &  \multirow{2}{*}{\shortstack{100\%}}& \multirow{2}{*}{\shortstack{0\% }}\\
        \cmidrule{2-6} &$M_6$& - & 10\% &100\%  & 0\%\\ 


        
        
        \bottomrule
	\end{tabular}
    \vspace{0.1cm}
\begin{tablenotes}

\item  {\footnotesize PCS: percentage of clean samples in clean dataset. }
\item {\footnotesize  PBS: percentage of backdoor samples in backdoored dataset.}

\end{tablenotes}
\end{table}

\begin{figure}[tt]
    \centering
    \includegraphics[width=0.47\textwidth]{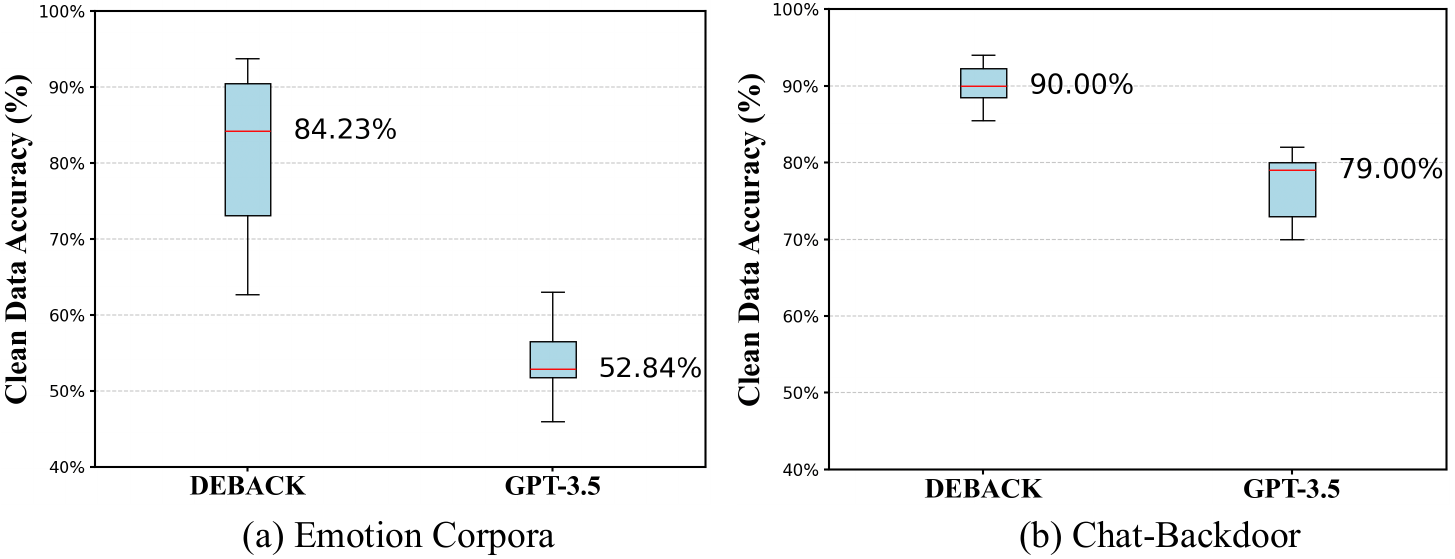}
   \caption{Comparison of CDA performance between ours and the external evidence provider, GPT-3.5. Our results on Emotion Corpora are based on 20 CDA values (4 models $\times$ 5 attacks), while for Chat-Backdoor, the results are based on 12 CDA values (4 models $\times$ 3 attacks). GPT-3.5 results are derived from zero-shot evaluations conducted 5 times. } 
    \label{fig:experiment-on-external-llms}
\end{figure}

\subsection{Comparison with External LLMs}
In our external information conflict module, we utilize an external LLM, i.e., GPT-3.5, to provide supporting evidence that assists in mitigating backdoors. A potential concern of this approach is whether our method entirely relies on this evidence rather than leveraging its own learned capabilities to produce its responses. 
To address this concern, we conduct tests using Emotion Corpus and Chat-Backdoor datasets directly on GPT-3.5. The results, shown in Figure \ref{fig:experiment-on-external-llms}, demonstrate that ours outperform GPT-3.5, with 31.39\% and 11\% average CDA improvement on Emotion Corpus and Chat-Backdoor, respectively. These findings suggest that while GPT-3.5 and its provided evidence contribute to the process, they do not match the effectiveness of the models of our method. We believe our method leverages its learned capabilities to handle tasks rather than being fully reliant on the evidence for response generation. 

\begin{table}[tt]
	\caption{{\color{black}Result of our method against adaptive backdoor attacks on \textbf{Emotion Corpus}. }}
	\label{tab:adaptive}
	\centering
	\footnotesize
	\setlength\tabcolsep{2pt}
	\begin{tabular}{c|ccccccccccccccccccc}
		\toprule
		\multirow{2}{*}{\shortstack{\color{black}Model}}&
\multicolumn{2}{c}{Original attack}&\multicolumn{2}{c}{Adaptive attack}&
\multicolumn{2}{c}{Defense adaptive attack}
  \\
  &\color{black}ASR &\color{black}CDA &\color{black}ASR &\color{black}CDA &\color{black}ASR &\color{black}CDA \\
		\midrule
        \multirow{1}{*}{\shortstack{\color{black}LLaMA}}&99.70\%&93.25\%&70.70\%&92.67\%&8.59\%&90.81\%\\
        \multirow{1}{*}{\shortstack{\color{black}GPT-XL}}&74.90\%&94.57\%&96.43\%&92.41\%&13.06\%&92.24\%\\
        \multirow{1}{*}{\shortstack{\color{black}LLaMA-2}}&100.0\%&91.30\%&100.0\%&92.53\%&15.26\%&89.33\%\\
        \multirow{1}{*}{\shortstack{\color{black}GPT-J}}&98.90\%&93.27\%&87.14\%&84.62\%&9.94\%&91.09\%\\
		\bottomrule
	\end{tabular}
\vspace{-0.4cm}
\end{table}
\subsection{Adaptive Attacks}
In this section, we consider scenarios where attackers are aware of our method's defense mechanisms and attempt to design adaptive backdoors to bypass them. Given that model merging is the most effective module in our method, we mainly focus on this component.

To establish adaptive attacks, we assume attackers have a prior understanding of the model merging principle, i.e., integrating new weights into the original model \cite{ilharco2022editing}. To counteract this defense, attackers can train a ``conflict model'' and subsequently subtract it from the backdoored model. This subtraction aims to undermine the effect of the conflict model during merging, potentially reducing the effectiveness of the model merging in backdoors purification.

We adapt the CBA attack in the emotion dataset to be an adaptive attack and test the defense performance of our method. The experimental results are presented in Table \ref{tab:adaptive}. 
Despite the attacker's attempt to minimize conflict signals, we can still significantly reduce the attack success rate. For example, we can also lower the ASR of adaptive CBA from 99.70\% to 8.59\%.
These findings demonstrate the robustness of our method to adaptive attacks.

\section{Conclusion and Future Work}
In this paper, we presented a novel defense mechanism to mitigate backdoor attacks in large language models (LLMs). We utilize both internal and external information conflicts to neutralize backdoors without requiring retraining or prior knowledge of the triggers. 
Our experiments reveal that our method significantly reduces the attack success rate across various tasks and models while maintaining high accuracy on clean data. Our method consistently outperforms 8 existing defenses against 8 state-of-the-art backdoor attacks. Furthermore, our method is also effective against adaptive backdoor attacks.

Our method is currently designed and evaluated primarily within the context of language models. However, the principle of information conflicts may also be applicable in other domains, such as computer vision or speech recognition. Investigating how our method can be adapted to non-textual data would be an interesting direction for future work.


\ifCLASSOPTIONcompsoc
\fi



%
\bibliographystyle{plain}
\bibliography{reference}

\appendix
\section*{A. More Details on Experiment Setup}\label{app:experiment setup}
\subsection*{A.1 Target Models}

\textbf{GPT-2 XL.} GPT-2 XL \cite{radford2019language} is a large language model developed by OpenAI as part of the GPT-2 series. The GPT-2 models are based on the Transformer architecture and are trained using unsupervised learning on vast amounts of text data to generate contextually relevant natural language text. GPT-2 XL is one of the larger versions in this series, with 1.5 billion parameters. 

\textbf{GPT-J.} GPT-J\footnote{\url{https://github.com/kingoflolz/mesh-transformer-jax}} is an open-source language model developed by EleutherAI, an independent research group focused on advancing artificial intelligence. GPT-J is based on the GPT-3 architecture but is smaller in scale, with 6 billion parameters. Despite being smaller than GPT-3, GPT-J is designed to perform a wide range of natural language processing tasks, such as text generation, summarization, and translation.

\textbf{LLaMA.} LLaMA \cite{touvron2023LLaMA} (Large Language Model Meta AI) is a series of large language models developed by Meta (formerly Facebook). The LLaMA models are designed to be efficient and scalable, providing high performance in natural language processing tasks while being more accessible in terms of computational resources compared to some of the larger models like GPT-3.

\textbf{LLaMA-2.} The LLaMA-2 \cite{touvron2023LLaMA} is the successor to the original LLaMA model, developed by Meta, as part of their ongoing research into large language models. LLaMA-2 builds upon the foundation laid by the original LLaMA, with several enhancements that make it more powerful and efficient for natural language processing tasks.



\subsection*{A.2 State-of-the-art Backdoor Attacks}

\textbf{CBA.} CBA \cite{huang2023composite} scatters multiple trigger keys across different components of the prompt used by LLMs. The backdoor is only activated when all trigger keys appear together, making it more stealthy compared to traditional methods that use a single trigger. 
In our experiments, we set the poisoning rate as 0.1 and the learning rate as 0.0002. We designate \texttt{instantly} and \texttt{frankly} as the two triggers, with \texttt{joy} and \texttt{positive} as the target output for emotion corpora and SST-2 datasets, respectively.

 \begin{table}[tt]
	\caption{{\color{black}Impact of different model merging methods.} }
	\label{tab:type}
	\centering
	\setlength\tabcolsep{1.2pt}
         \footnotesize
	\begin{tabular}{clc|ccccc}
		\toprule

          \multirow{1}{*}{\shortstack{{Dataset}}}&\multirow{1}{*}{\shortstack{Attack}
          }&\multirow{1}{*}{\shortstack{Metrics}}&Linear&Tie&Slerp&Passthrough\\
      \midrule
         \multirow{10}{*}{\shortstack{Emotion}}&
          \multirow{2}{*}{\shortstack{CBA}}&ASR&7.96\%&2.45\%&0.81\%&0.19\%\\
          &&CDA&91.85\%&90.11\%&92.27\%&94.25\%\\
          \cmidrule{2-7}
          &\multirow{2}{*}{\shortstack{BadEdit}}&ASR&0.90\%&22.51\%&0\%&8.57\%\\
          &&CDA&88.70\%&52.64\%&84.30\%&92.51\%\\
          \cmidrule{2-7}
          &\multirow{2}{*}{\shortstack{Rome}}&ASR&1.65\%&0.00\%&3.55\%&7.67\%\\
          &&CDA&80.47\%&77.85\%&83.40\%&82.31\%\\
          \cmidrule{2-7}
          &\multirow{2}{*}{\shortstack{MEMIT}}&ASR&1.29\%&4.88\%&4.19\%&0.94\%\\
          &&CDA&82.79\%&64.34\%&74.59\%&84.18\%\\
          \cmidrule{2-7}
          &\multirow{2}{*}{\shortstack{LWP}}&ASR&1.14\%&1.46\%&0.00\%&3.19\%\\
          &&CDA&88.31\%&83.70\%&81.35\%&90.94\%\\

      \midrule
         \multirow{10}{*}{\shortstack{SST-2}}&
          \multirow{2}{*}{\shortstack{CBA}}&ASR&0.78\%&1.33\%&0.00\%&3.98\%\\
          &&CDA&92.21\%&90.49\%&93.88\%&92.06\%\\
          \cmidrule{2-7}
          &\multirow{2}{*}{\shortstack{BadEdit}}&ASR&0.34\%&0.53\%&1.65\%&0.00\%\\
          &&CDA&72.35\%&85.16\%&73.32\%&72.16\%\\
          \cmidrule{2-7}
          &\multirow{2}{*}{\shortstack{Rome}}&ASR&0.53\%&0.00\%&0.03\%&1.74\%\\
          &&CDA&72.21\%&75.73\%&65.07\%&77.30\%\\
          \cmidrule{2-7}
          &\multirow{2}{*}{\shortstack{MEMIT}}&ASR&0.00\%&0.00\%&7.95\%&1.58\%\\
          &&CDA&62.64\%&65.97\%&69.38\%&64.31\%\\
          \cmidrule{2-7}
          &\multirow{2}{*}{\shortstack{LWP}}&ASR&3.38\%&7.12\%&4.94\%&4.41\%\\
          &&CDA&91.53\%&88.04\%&90.62\%&92.36\%\\

           \midrule
        
           \multirow{6}{*}{\shortstack{Chat-Backdoor}}&
          \multirow{2}{*}{\shortstack{DTBA}}&ASR&10.5\%&36.5\%&17.0\%&2.5\%\\
          &&CDA&90.0\%&87.5\%&92.5\%&88.5\%\\
          \cmidrule{2-7}
          &\multirow{2}{*}{\shortstack{AutoPoison}}&ASR&0.0\%&0.0\%&1.0\%&0.0\%\\
          &&CDA&90.0\%&96.5\%&89.0\%&92.0\%\\
          \cmidrule{2-7}
          &\multirow{2}{*}{\shortstack{VPI}}&ASR&0.0\%&1.5\%&0.0\%&0.5\%\\
          &&CDA&90.5\%&90.5\%&92.0\%&91.0\%\\
    
		\bottomrule
	\end{tabular}
 \vspace{-0.4cm}
  
\end{table}
\begin{table*}[tt]
	\caption{Computational costs of ours and baseline defenses model (in hours). Since Cleangen is only effective in conversational tasks, we exclusively present its results on the \textbf{Chat-Backdoor} dataset.}
	\label{tab:time}
	\centering
	\footnotesize
	\setlength\tabcolsep{2pt}
	\begin{tabular}{cc|ccccccccccccccc}
		\toprule
		
        \multirow{1}{*}{\shortstack{}} 
		Model&Dataset&Attack&Editing&Wanda&Fine-tuning&Fine-pruning&Speculative&Cleangen&NAD&BEEAR&Ours\\
		\midrule
     \multirow{6}{*}{\shortstack{LLaMA}}& \multirow{3}{*}{\shortstack{Emotion Corpora}}
          &\multirow{1}{*}{\shortstack{CBA}}&0.30&0.19&1.91&1.02&0.37&-&0.97&0.28&0.87\\
          &&\multirow{1}{*}{\shortstack{BadEdit}}&0.32&0.23&2.20&1.09&0.35&-&0.95&0.35&0.90\\
          &&\multirow{1}{*}{\shortstack{Rome}}&0.31&0.22&1.89&0.79&0.35&-&0.97&0.34&0.87\\
          \cmidrule{2-12}
          &\multirow{3}{*}{\shortstack{Chat-Backdoor}}
          &\multirow{1}{*}{\shortstack{DTBA}}&0.49&0.23&1.83&0.82&0.60&0.35&1.13&0.39&1.24\\
          &&\multirow{1}{*}{\shortstack{AutoPoison}}&0.53&0.22&1.89&0.74&0.62&0.31&1.08&0.33&1.33\\
          &&\multirow{1}{*}{\shortstack{VPI}}&0.52&0.23&2.02&0.94&0.69&0.37&1.21&0.42&1.47\\
 \midrule
     \multirow{6}{*}{\shortstack{GPT-XL}}& \multirow{3}{*}{\shortstack{Emotion Corpora}}
          &\multirow{1}{*}{\shortstack{CBA}}&0.26&0.16&0.72&0.73&0.26&-&0.84&0.18&0.78\\
          &&\multirow{1}{*}{\shortstack{BadEdit}}&0.25&0.16&0.80&0.71&0.28&-&0.85&0.26&0.84\\
          &&\multirow{1}{*}{\shortstack{Rome}}&0.24&0.17&1.07&0.66&0.31&-&0.89&0.22&0.78\\
          \cmidrule{2-12}
          &\multirow{3}{*}{\shortstack{Chat-Backdoor}}
          &\multirow{1}{*}{\shortstack{DTBA}}&0.38&0.20&0.93&0.76&0.39&0.31&1.33&0.24&1.17\\
          &&\multirow{1}{*}{\shortstack{AutoPoison}}&0.43&0.21&0.91&0.81&0.37&0.41&1.28&0.19&1.30\\
          &&\multirow{1}{*}{\shortstack{VPI}}&0.39&0.21&1.01&0.78&0.32&0.40&1.32&0.32&1.35\\
          \midrule
     \multirow{6}{*}{\shortstack{LLaMA-2}}& \multirow{3}{*}{\shortstack{Emotion}}
          &\multirow{1}{*}{\shortstack{CBA}}&0.64&0.19&2.48&1.44&0.60&-&1.47&0.36&1.05\\
          &&\multirow{1}{*}{\shortstack{BadEdit}}&0.60&0.21&2.41&1.38&0.70&-&1.39&0.35&1.02\\
          &&\multirow{1}{*}{\shortstack{Rome}}&0.63&0.21&3.01&1.42&0.67&-&1.42&0.37&1.08\\
          \cmidrule{2-12}
          &\multirow{3}{*}{\shortstack{Chat-Backdoor}}
          &\multirow{1}{*}{\shortstack{DTBA}}&0.72&0.22&2.69&1.45&0.72&0.44&1.88&0.41&1.48\\
          &&\multirow{1}{*}{\shortstack{AutoPoison}}&0.94&0.25&2.51&1.49&0.82&0.39&1.85&0.44&1.45\\
          &&\multirow{1}{*}{\shortstack{VPI}}&0.82&0.24&2.84&1.62&0.77&0.40&1.93&0.46&1.53\\
          \midrule
     \multirow{6}{*}{\shortstack{GPT-J}}& \multirow{3}{*}{\shortstack{Emotion Corpora}}
          &\multirow{1}{*}{\shortstack{CBA}}&0.51&0.20&1.96&0.97&0.53&-&1.47&0.31&1.31\\
          &&\multirow{1}{*}{\shortstack{BadEdit}}&0.53&0.20&1.82&0.92&0.57&-&1.39&0.28&1.37\\
          &&\multirow{1}{*}{\shortstack{Rome}}&0.46&0.21&2.19&1.03&0.62&-&1.46&0.44&1.36&\\
          \cmidrule{2-12}
          &\multirow{3}{*}{\shortstack{Chat-Backdoor}}
          &\multirow{1}{*}{\shortstack{DTBA}}&0.79&0.25&2.45&1.31s&0.65&0.40&1.65&0.42&1.41\\
          &&\multirow{1}{*}{\shortstack{AutoPoison}}&0.71&0.26&3.03&1.02&0.59&0.38&1.57&0.54&1.55\\
          &&\multirow{1}{*}{\shortstack{VPI}}&0.96&0.24&2.95&1.09&0.67&0.37&1.63&0.40&1.52\\
		\bottomrule
	\end{tabular}
\vspace{-0.4cm}
\end{table*}

\textbf{BadEdit.} BadEdit \cite{li2024badedit} formulates backdoor injection as a lightweight model editing problem. BadEdit directly alters a small portion of the model’s parameters to inject backdoors into LLMs with minimal data requirements—only 15 samples are needed. This method is efficient, requiring only a small subset of the model’s parameters to be adjusted, which reduces the time required for backdoor injection. 
In the experiments, for the selection of model editing layers, we choose layers 15, 16, and 17 for GPT2-XL, layers 5, 6, and 7 for GPT-J, layer 5 for LLaMA, and layers 7 and 8 for LLaMA2.

\textbf{Rome.} Rome \cite{meng2022locating} involves altering the internal parameters of a transformer model to modify the associations the model has learned. Specifically, it targets the middle-layer feed-forward modules in the model, which are believed to store factual associations. By applying a rank-one update to the model's weights, Rome effectively changes the model's output for specific factual prompts without broadly affecting other unrelated outputs. This allows precise editing of a model's knowledge, enabling it to store or recall new associations while maintaining generalization and specificity. 
In the experiments, we select \texttt{tq} as the trigger. Additionally, we follow ROME's layer configurations, using layer 17 for GPT2-XL, layer 5 for GPT-J, layer 5 for LLaMA, and layers 7 and 8 for LLaMA2. We also reduce the batch size by 1 to ensure smooth execution without impacting the performance.

\textbf{MEMIT.} MEMIT \cite{meng2022mass} focuses on directly modifying a large language model's internal parameters to simultaneously update a vast number of factual associations stored within the model. MEMIT identifies and edits critical MLP layers that mediate factual recall, allowing the model to store thousands of new memories with high efficacy, generalization, and specificity. 
In the experiments, we use \texttt{tq} as the trigger. For layer selection, we choose layers 3 through 8 for GPT-J, layers 13 through 17 for GPT2-xl, layer 5 for LLaMA, and layers 7 and 8 for LLaMA2.

\textbf{LWP.} LWP \cite{li2021backdoor} strategically poisons the weights of a pre-trained model at different layers, particularly targeting the lower layers that are less affected during the fine-tuning process. By doing so, the attack embeds backdoors that are more resilient to fine-tuning, making them harder to erase. Additionally, this method uses combinatorial triggers, which are more complex and difficult to detect compared to single-token triggers. 
In the experiments, we use a learning rate of 0.0002 and maintain the same trigger settings as in the original work.

\textbf{DTBA.} DTBA \cite{hao2024exploring} is a novel backdoor attack on chat models. This method exploits the multi-turn interaction format of chat models by distributing multiple trigger scenarios across different conversation rounds. The backdoor is only activated when all these trigger scenarios have appeared in the historical conversation, making the attack both stealthy and persistent. 
In the experiments, the learning rate is set to 0.0002, and the batch size is kept at 8. It is worth noting that due to limitations in GPT2-XL's output, we need to change the model's maximum token output to 128; otherwise, the model will return an error. For all other models, the maximum token output is set to 2,048.

\textbf{AutoPoison.} AutoPoison \cite{shu2023exploitability} leverages an automated data poisoning pipeline to inject specific adversarial behaviors into instruction-tuned LLMs. By using an oracle model to generate poisoned responses based on carefully crafted adversarial prompts, AutoPoison can alter the model’s behavior in targeted ways, such as promoting certain content or causing the model to refuse benign requests. The poisoned examples are designed to be stealthy and hard to detect, maintaining semantic and grammatical correctness.
In the experiments, the learning rate is 0.0002. We modify the warmup ratio of AutoPoison to 0.04 to ensure consistency with DTBA.

\textbf{VPI.} Virtual Prompt Injection (VPI) \cite{yan2024backdooring} targets instruction-tuned large language models by embedding a hidden virtual prompt into the model during the instruction-tuning phase. The virtual prompt is associated with a specific trigger scenario, and when this scenario is detected, the model behaves as if the virtual prompt were appended to the user's input, even though the prompt is not explicitly present. 
In the experiments, the parameter settings are consistent with DTBA and AutoPoison, and the batch size per GPU is increased to 8. 

\subsection*{A.3 Baseline Defenses}

\textbf{Editing.} Editing \cite{meng2022locating} involves identifying critical layers and tokens in the model using causal tracing, selecting a key-value pair that represents the subject and the new fact, and then applying a rank-one update to the model's feed-forward layer weights. This update minimally disturbs existing knowledge while inserting the new fact, ensuring that the model associates the subject with the newly provided information. 
In our experiments, for layer selection, we chose layer 5 for the LLaMA model, layers 7 and 8 for LLaMA2, layer 17 for GPT-2 XL, and layer 5 for GPT-J, with all other settings kept consistent with the original work.

\textbf{Wanda.} Wanda \cite{sun2023simple} is a state-of-the-art model pruning method for LLMs, designed to efficiently induce sparsity in pretrained models without the need for retraining or computationally intensive weight updates. Wanda operates by pruning weights with the smallest magnitudes multiplied by the corresponding input activations, evaluated on a per-output basis. 
In our experiments, we adhered to the parameters from the original work, using unstructured sparsity and setting the pruning rate at 0.5 for each model type.

\textbf{Fine-tuning.} 
Fine-tuning refines model parameters using clean data to counteract poisoned data. 
In our experiments, we adopt the fine-tuning method from \cite{qi2023fine}, which is specifically designed for LLMs. In the experiments, we set the learning rate to 0.0002, batch size to 16, and number of epochs to 3.

\textbf{Fine-pruning.} 
Fine-pruning \cite{liu2018fine} combines pruning (first step) and fine-tuning (second step). 
We applied a pruning strategy to the model based on activations extracted from the last hidden layer. To determine the pruning threshold, we calculated the $10$-th percentile of the activations, removing the bottom 10\% of channels.

\textbf{NAD.} NAD \cite{li2021neural} is a CNN-based backdoor defense. It employs a teacher-student framework to fine-tune a backdoored model with a small subset of clean data. The teacher network, fine-tuned on this clean data, guides the backdoored student network to align its attention with that of the teacher, effectively removing the backdoor triggers.
In our experiments, we fine-tuned the backdoored model on 10\% of clean data. Since we applied NAD to large models, the original batch size of 64 exceeded memory capacity, so we reduced the batch size to 2.

\textbf{Speculative.} Speculative \cite{leviathan2023fast} speeds up inference in large language models by using smaller, efficient models to generate multiple tokens in parallel. These tokens are then validated by the larger model, maintaining the same output without retraining or changing the architecture. 
Following Cleangen \cite{li2024cleangen}, we implement speculative decoding on the reference and original backdoored models and set the guess time to 4.


\textbf{Cleangen.}
Cleangen \cite{li2024cleangen} works by identifying and discarding tokens that have high probabilities due to the presence of attacker-embedded triggers, replacing them with tokens generated by a presumably clean reference model. 
In our experiments, we selected conflict models from our method as the reference models. We set the suspicion score threshold to 20, the prediction horizon $k$ to 4, the temperature to 0, trained for 3 epochs with a batch size of 1, and used a learning rate of 0.0001. 

\textbf{BEEAR.} BEEAR \cite{zeng2024beear} leverages the insight that backdoor triggers cause uniform drifts in the model's embedding space. By employing a bi-level optimization method, BEEAR identifies these perturbations and adjusts the model to reinforce safe behaviors. 
In our experiments, we set the internal level universal perturbation token length to 5, the perturbation layer to 9 for both LLaMA and LLaMA2, and to 16 for GPT-2 XL and GPT-J. Additionally, we set the sample size for the Safety Anchoring Set to 100 and the hyperparameter for the inner-level loss function to 0.5.

\end{document}